\documentclass[manuscript,screen,authorversion]{acmart}

\setcopyright{none}
\copyrightyear{2021}
\acmYear{2021}

 \acmConference[]{}{}{}
 \acmBooktitle{}

\usepackage{graphicx}
\usepackage{url}
\usepackage{xcolor}
\usepackage{algorithm}
\usepackage{array}
\usepackage{multirow}
\usepackage[noend]{algpseudocode}
\usepackage[acronym,nohypertypes={acronym,notation}]{glossaries}
\usepackage{enumitem}
\usepackage{soul}
\usepackage[tight]{subfigure}

\newacronym{nn}{NN}{Neural Network}
\newacronym{ann}{ANN}{Artificial Neural Network}
\newacronym{annpl}{ANNs}{Artificial Neural Networks}
\newacronym{cnn}{CNN}{Convolutional Neural Network}
\newacronym{cnnpl}{CNNs}{Convolutional Neural Networks}
\newacronym{gnn}{GNN}{Graph Neural Network}
\newacronym{dnn}{DNN}{Deep Neural Network}
\newacronym{gnnpl}{GNNs}{Graph Neural Networks}
\newacronym{rnn}{RNN}{Recursive Neural Network}
\newacronym{rnnpl}{RNNs}{Recursive Neural Networks}
\newacronym{kg}{KG}{Knowledge Graph}
\newacronym{fpga}{FPGA}{Field Programmable Gate Array}
\newacronym{gpu}{GPU}{Graphics Processing Unit}
\newacronym{cpu}{CPU}{Central Processing Unit}
\newacronym{dl}{DL}{Deep Learning}
\newacronym{gcn}{GCN}{Graph Convolutional Networks}
\newacronym{spmm}{SpMM}{sparse-dense matrix multiplications}
\newacronym{asic}{ASIC}{Application-Specific Integrated Circuits}
\newacronym{bptt}{BPTT}{Back-Propagation-Through-Time}
\newacronym{gk}{GK}{Graph Kernels}
\newacronym{ml}{ML}{Machine Learning}
\newacronym{gat}{GAT}{Graph Attention Networks}
\newacronym{nlp}{NLP}{Natural Language Processing}
\newacronym{gin}{GIN}{Graph Isomorphism Network}
\newacronym{gae}{GAE}{Graph Autoencoders}
\newacronym{gan}{GAN}{Generative Adversarial Networks}
\newacronym{mpnn}{MPNN}{Message Passing Neural Network}
\newacronym{nlnn}{NLNN}{Non-local Neural Network}
\newacronym{cp}{CP}{control plane}
\newacronym{dp}{DP}{data plane}
\newacronym{pe}{PE}{processing element}
\newacronym{noc}{NoC}{Network-on-Chip}

\begin{document}

\title{Computing Graph Neural Networks: A Survey from Algorithms to Accelerators}

\author{Sergi Abadal}
\email{abadal@ac.upc.edu}
\orcid{0000-0003-0941-0260}

\author{Akshay Jain}
\email{akshay.jain@upc.edu}
\orcid{0000-0002-6986-2720}

\author{Robert Guirado}
\email{rguiado@ac.upc.edu}
\orcid{0000-0002-3559-932X}

\author{Jorge L\'{o}pez-Alonso}
\email{jorlopezalonso@gmail.com}

\author{Eduard Alarc\'{o}n}
\email{eduard.alarcon@upc.edu}
\orcid{0000-0001-7663-7153}

\affiliation{%
  \institution{Universitat Polit\`{e}cnica de Catalunya}
  \streetaddress{C/ Jordi Girona 1-3}
  \city{Barcelona}
  \country{Spain}
  \postcode{08034}
}


\renewcommand{\shortauthors}{S. Abadal, A. Jain, R. Guirado, J. L\'{o}pez-Alonso, and E.  Alarc\'{o}n}
\renewcommand{\hl}[1]{#1}

\begin{abstract}
\acrfull{gnnpl} have exploded onto the machine learning scene in recent years owing to their capability to model and learn from graph-structured data. 
Such an ability has strong implications in a wide variety of fields whose data is inherently relational, for which conventional neural networks do not perform well.
Indeed, as recent reviews can attest, research in the area of GNNs has grown rapidly and has lead to the development of a variety of GNN algorithm variants as well as to the exploration of groundbreaking applications in chemistry, neurology, electronics, or communication networks, among others.
At the current stage of research, however, the efficient processing of GNNs is still an open challenge for several reasons. Besides of their novelty, GNNs are hard to compute due to their dependence on the input graph, their combination of dense and very sparse operations, or the need to scale to huge graphs in some applications.
In this context, this paper aims to make two main contributions. On the one hand, a review of the field of GNNs is presented from the perspective of computing. This includes a brief tutorial on the GNN fundamentals, an overview of the evolution of the field in the last decade, and a summary of operations carried out in the multiple phases of different GNN algorithm variants. On the other hand, an in-depth analysis of current software and hardware acceleration schemes is provided, from which a hardware-software, graph-aware, and communication-centric vision for GNN accelerators is distilled. 
\end{abstract}



\maketitle

\section{Introduction} 
\label{intro}
Machine Learning (ML) has taken the world by storm and has become a fundamental pillar of engineering due to its capacity to solve extremely complex problems, to detect intricate features in oceans of data, or to automatically generate alternatives that outperform well-engineered, well-known, carefully optimized solutions. As a result, the last decade has witnessed an explosive growth in the use of Deep Neural Networks (DNNs) in pursuit of exploiting the advantages of ML in virtually every aspect of our lives \cite{Lecun2015}: computer vision \cite{resnet}, natural language processing \cite{nlp}, medicine \cite{Esteva2017} or economics \cite{stock} are just a few examples.


However, and in spite of its all-pervasive applicability and potential, it is well-known that not all neural network architectures fit to all problems \cite{Battaglia2018}. DNNs take the input data and attempt to extract knowledge taking into account the inductive bias that the connection architecture of the DNN generates. This, in essence, means that the number of DNN layers and their pre-assumed connections determines its suitability to certain tasks. For instance, by not making any assumption on the structure of the data, conventional fully-connected neural networks are able to master a wide range of tasks at the cost of being less efficient in general than other DNNs \cite{Botvinick2019}. In contrast, techniques such as \acrfull{cnnpl} or \acrfull{rnnpl} are biased towards extracting knowledge from the locality and temporal sequentiality of data. This makes them a better fit for specific tasks such as image recognition or treatment of temporal signals, yet incapable of efficiently handling data with arbitrary structures \cite{wang2016cnn}.

In light of the above, there has been a recent interest in deep learning techniques able to model graph-structured data \cite{Bronstein2017,Zhang2020a,Battaglia2018, vectornet, Xugeng2019,uber}. This structure is inherent to a plethora of problems in the field of complex systems in general, and applicable to particular fields such as communication networks where the topology and routing decisions determine its performance \cite{Rusek2019}, synthetic chemistry where molecular structures determine the compound properties \cite{Gilmer2017}, social networks where emergent behavior can arise through personal relations \cite{Nettleton2013}, or neuroscience where specific connections between neuron types and brain areas determine brain function \cite{BrainGNNIB}, among many others.


\acrfull{gnnpl} are a set of connectivity-driven models that, since the late 2000s, have been addressing the need for geometric deep learning \cite{Gori2005, Scarselli2009}. 
In essence, \acrshort{gnnpl} adapt their structure to that of an input graph and, through an iterative process of aggregation of information across vertices, capture the complex dependences of the underlying system. This allows to predict properties for specific nodes, connections, or the graph as a whole, and generalize to unseen graphs. 
Due to these powerful features, many relevant applications such as molecule property prediction \cite{Fout2017}, recommender systems \cite{Fan2019}, natural language processing \cite{nlp}, traffic speed prediction \cite{Xie2020}, critical data classification \cite{YongSccar}, computer vision \cite{Wang2018}, particle physics \cite{ju2020graph}, resource allocation in computer networks \cite{Rusek2019a}, already utilize \acrshort{gnnpl} to accomplish their tasks.  

\begin{figure}[!t]
    \centering
    \includegraphics[width=0.6\columnwidth]{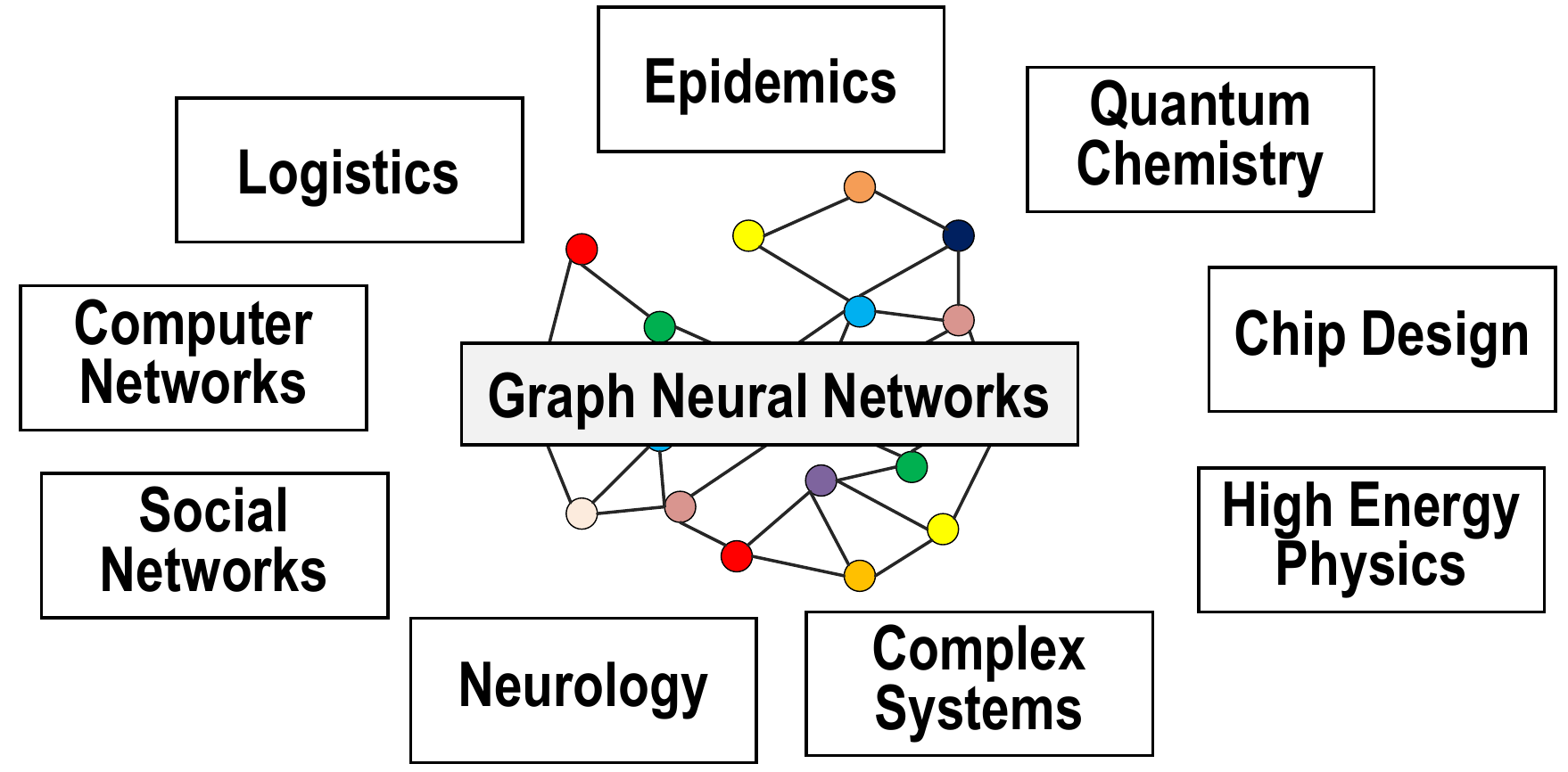}
    \vspace{-0.2cm}
    \caption{Graph Neural Networks (GNNs) as enablers of a plethora of applications in fields that hinge on graph-structured data.}
    \vspace{-0.6cm}
    \label{Fig1}
\end{figure}

For all these reasons, recent years have seen a rapid increase in research activity in the field of GNNs (see Fig. \ref{fig:number_of_papers_evolution}). Intense efforts are being directed towards improving the efficiency of algorithms, especially for large graphs, and towards demonstrating their efficacy for the aforementioned application areas. The interested reader will find multiple reviews of the state of the art in GNN algorithms and applications in the literature \cite{Battaglia2018,Zhou2018a,wu2019comprehensive,chami2020machine,Lamb2020,Bronstein2017,Hamilton2017,Zhang2020a}, most of which we briefly analyze in Table \ref{tab1}. Other key aspects relevant or adjacent to GNNs such as network embedding \cite{Cui2019b}, graph attention models \cite{Lee2019}, or network structure inference \cite{Brugere2018} have also received a comprehensive review.


\begin{table*}
    \centering
   \renewcommand{\arraystretch}{1.5}
   \small
    \caption{Background literature: surveys about GNNs (first block) and including GNNs (second block)}
    \vspace{-0.4cm}
    \begin{tabular}{|m{4.2cm}|>{\centering\arraybackslash}p{10.2cm}|} \hline
         \centering \textbf{Study [Reference] (Year)} & \textbf{Contributions} \\ \hline\hline

        \begin{minipage}[t]{\linewidth}
        Relational Inductive Biases, Deep Learning, and Graph Networks \cite{Battaglia2018} (2018)  
        \end{minipage}
         & \begin{minipage}[t]{\linewidth}
        \begin{itemize} [nosep,after=\strut,leftmargin=4.5pt]
            \item Presents the idea of a graph network as a generalization of GNNs with building blocks
            \item Encompasses well-known models, such as fully connected, convolutional and recurrent networks.  
        \end{itemize} 
        \end{minipage}\\ \hline
        
        \begin{minipage}[t]{\linewidth}
        Graph Neural Networks: A Review of Methods and Applications \cite{Zhou2018a} (2018)
         \end{minipage}
         & \begin{minipage}[t]{\linewidth}
        \begin{itemize} [nosep,after=\strut,leftmargin=4.5pt]
            \item Presents a survey of the various \acrshort{gnn} models
            \item Discusses the applications where \acrshort{gnnpl} can be utilized and provides a taxonomy
            \item Proposes open research problems, such as dynamicity and scalability in \acrshort{gnnpl}  
        \end{itemize} 
        \end{minipage}\\ \hline

        \begin{minipage}[t]{\linewidth}
        A Comprehensive Survey on Graph Neural Networks \cite{wu2019comprehensive}  (2021)
         \end{minipage}& 
         \begin{minipage}[t]{\linewidth}
        \begin{itemize} [nosep,after=\strut,leftmargin=4.5pt]
            \item Overviews of \acrshort{gnnpl} in data mining and machine learning areas  
            \item Provisions a taxonomy for \acrshort{gnn} models
            \item Details the application areas of \acrshort{gnnpl}
            \item Presents potential research directions, such as in scalability, dynamicity of \acrshort{gnnpl}, etc.
        \end{itemize} 
        \end{minipage}\\ \hline

        \begin{minipage}[t]{\linewidth}
        Deep Learning on Graphs: A Survey \cite{Zhang2020a}  (2020)
         \end{minipage} 
         & \begin{minipage}[t]{\linewidth}
        \begin{itemize} [nosep,after=\strut,leftmargin=4.5pt]
            \item Provides a discussion on graph versions of recurrent and convolutional networks, autoencoders, reinforcement-learning and adversarial methods
            \item Presents the application areas and future research directions for deep learning on graphs%
        \end{itemize}
        \end{minipage}\\ \hline
        
        \begin{minipage}[t]{\linewidth}
         \hl{Machine Learning on Graphs: A Model and Comprehensive Taxonomy} \cite{chami2020machine} (2020)%
         \end{minipage} &  \begin{minipage}[t]{\linewidth}
        \begin{itemize} [nosep,after=\strut,leftmargin=4.5pt]
            \item \hl{Presents a taxonomy to classify graph learning methods, from graph embeddings to GNNs}
            \item \hl{Proposes an encoder-decoder model that unifies all methods in a single approach}
            \item \hl{Expresses 30+ graph learning techniques using the proposed model}
        \end{itemize}
        \end{minipage}\\ \hline

        \begin{minipage}[t]{\linewidth}
         Graph Neural Networks Meet Neural-Symbolic Computing: A Survey and Perspective \cite{Lamb2020} (2020)
         \end{minipage} & \begin{minipage}[t]{\linewidth}
        \begin{itemize} [nosep,after=\strut,leftmargin=4.5pt]
            \item Elaborates the relationship between \acrshort{gnnpl} and Neural-Symbolic Computing%
            \item Develops multiple \acrshort{gnn} models with the perspective of being applied to Neural-Symbolic computing%
        \end{itemize}
        \end{minipage}\\ \hline\hline

        \begin{minipage}[t]{\linewidth}
        \begin{itemize} [nosep,after=\strut,leftmargin=0pt]
         \vspace{0.01mm}
         \item[] Geometric Deep Learning: Going beyond Euclidean data \cite{Bronstein2017}  (2017)
         \end{itemize}
         \end{minipage} &  \begin{minipage}[t]{\linewidth}
        \begin{itemize} [nosep,after=\strut,leftmargin=4.5pt]
            \item Proposes Geometric Deep Learning as an umbrella term for models that operate on non-euclidean dataset representations, including GNNs.
            \item Within GNNs, provides a thorough review of convolutional models
        \end{itemize}
        \end{minipage}\\ \hline
        
        \begin{minipage}[t]{\linewidth}
         Representation Learning on Graphs: Methods and Applications \cite{Hamilton2017} (2017)%
         \end{minipage} &  \begin{minipage}[t]{\linewidth}
        \begin{itemize} [nosep,after=\strut,leftmargin=4.5pt]
            \item Reviews the advancements in the area of representation learning on graphs
            \item Primary focus is on the network embedding methods
        \end{itemize}
        \end{minipage}\\ \hline

     \end{tabular}
     \vspace{-0.3cm}
    \label{tab1}
\end{table*}

As we will see along this paper, however, less attention has been placed on the efficient processing of such new type of neural networks. While the issue has already been investigated in significant depth for \acrshort{cnnpl} or \acrshort{rnnpl} \cite{Chen2017,Mittal2020, maeri, shi, ucnn, eyeriss2}, GNN processing remains largely unexplored. This is because GNNs are relatively novel and pose unique computing challenges, including the need to (i) support both dense and extremely sparse operations, (ii) adapt the computation to the specific GNN algorithm variant and the structure of the graph at hand, and (iii) scale to very large graphs to realize its potential in certain applications. \hl{Even though advances in sparse/irregular tensor processing} \cite{dave2020hardware} \hl{and graph processing} \cite{ham2016graphicionado, wang2016gunrock} \hl{may prove useful in accelerating GNNs, addressing their unique computing challenges requires more specialized proposals. Some attempts have been done} from a software perspective, i.e. adapting the GNN operations to better match the capabilities of CPUs or GPUs \cite{Tripathy2020, Ma2019, gnnadvisor}; and from a hardware perspective, i.e. designing custom processors tailored to the demands of GNNs \cite{EnGN, Yan2020, AWB, Auten2020}. However, recent surveys and reviews \cite{Battaglia2018, Zhou2018a, wu2019comprehensive, chami2020machine, Lamb2020, Bronstein2017, Hamilton2017, Zhang2020a} lack of comprehensive analysis of such advances. 

\begin{figure*}[!t]
    \centering
    \includegraphics[width=\textwidth]{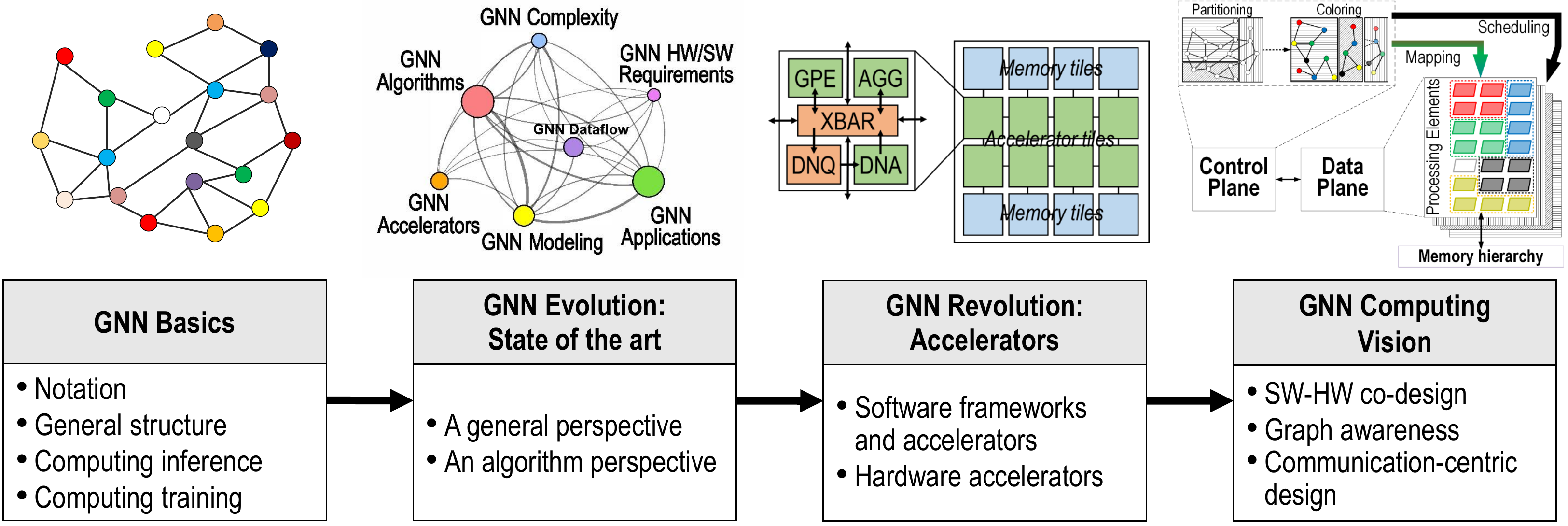}
    \vspace{-0.5cm}
    \caption{Graphical abstract of this survey from the GNN fundamentals \hl{(Section }\ref{sec2}) to the proposed architectural vision \hl{(Section }\ref{sec5}).}
    \vspace{-0.4cm}
    \label{Fig2}
\end{figure*}

This paper aims to bridge this gap by presenting, for the first time, a review of the field of GNNs from the perspective of computing. To that end, we make the following contributions as summarized \hl{in} Fig. \ref{Fig2}: we start by providing a comprehensive and tutorial-like description of the fundamentals of \acrshort{gnnpl}, trying to unify notation. Then, using a Knowledge Graph (KG) approach, we chart the evolution of the field from its inception to the time of this writing, delving into the duality between GNN algorithms (seeing them as learning systems) and their associated computation (seeing them as sets of matrix multiplications and non-linear operations). From that analysis, we identify GNN computing as a nascent field. We finally focus on the computation aspect and provide an in-depth analysis of current software and hardware acceleration schemes, from which we also outline new potential research lines in GNN computing. To the best of the authors' knowledge, this is the first work providing a thorough review of GNN research from the perspective of computation, charting the evolution of the research area and analyzing existing libraries and accelerators.

The rest of this paper is organized as follows: In Section \ref{sec2}, we discuss the basics of the \acrshort{gnnpl}. Section \ref{sec3} presents the evolution of the research area from multiple perspectives. In Section \ref{sec4}, we expose the emergent area of \acrshort{gnn} accelerators, summarizing recent works and elaborating upon the existing challenges and opportunities. Next, in Section \ref{sec5}, we present our vision for the architectural design of \acrshort{gnn} accelerators with a focus on internal communication requirements. We conclude this paper in Section \ref{conclusion}.

\section{Fundamentals of Graph Neural Networks} \label{sec2}
In this section, we discuss the basics of \acrshort{gnnpl} through a description of their building blocks and their role during the computation, both in inference and training.     

\subsection{Notation} \label{sec2a}
We first describe the main notation for \acrshort{gnnpl} as summarized in Table \ref{tab:notation}. Let a graph $G=(V,E)$ be defined by a set of vertices $V$, and a set of edges $E$ that connect some of the vertices in $V$ together. In particular, each vertex $v\in V$ has a neighbourhood set $N(v)$ determined by the edges connecting it to other vertices \hl{or the sampling set imposed by the GNN algorithm}. Further, each vertex $v$ contains a vertex feature representation $h_{v}$, and each edge $e\in E$  contains an edge feature representation $g_{e}$. The vertex or edge feature representations are generally one-dimensional vectors containing the scalar attributes that define them. Similarly, the graph may be associated to \hl{a} global feature representation $y$ containing graph-wide attributes. 
For example, in a social networking graph, vertices might be users with attributes such as encoded name or location, whereas the edges might be the interaction between two users such as comments/likes on a picture. Graph-wide features may be the number of users living a certain area or voting a certain political party.

GNNs essentially calculate a set of output feature representations for the vertices $h_v$, edges $g_{e}$, and complete graph $y$, respectively. Following with the example above, for targeting ads in a social network, output features of a vertex could be the probability of being interested in cars. It can thus be observed that, as in any other neural network, the dimensionality of the output feature vectors will be generally different than that of the input.

\begin{table}[!htb]
    \centering
    \caption{Graph representation notations}
    \vspace{-0.4cm}
    \begin{tabular}{|c|c||c|c|} \hline
         $V$ & Set of vertices of the graph & $h_{v},h_{v}^{(l)},h_{v}^{L}$ & Input, hidden, output feature vector of vertex $v$ \\  \hline
         $E$ & Set of edges of the graph & $g_{e},g_{e}^{(l)},g_{e}^{L}$ & Input, hidden, output feature vector of edge $e$ \\ \hline
         $N(v)$ & Set of neigbours of vertex $v$ & $\rho_{V}^{(l)}$, $\rho_{E}^{(l)}$ & Node and edge aggregation functions of layer $l$ \\ \hline
         $L$ & Number of GNN layers & $\phi_{V}^{(l)}$, $\phi_{E}^{(l)}$ & Node and edge combination functions of layer $l$ \\ \hline
         $y$ & Output global vector & $W^{(l)}_{V}$, $W^{(l)}_{E}$ & Node and edge weight matrices of layer $l$ \\ \hline     
         
    \end{tabular}
    \vspace{-0.2cm}
    \label{tab:notation}
\end{table}


As we will see in Section \ref{sec:structure}, a GNN is divided in multiple layers. In each layer $l \in [1,L]$, there is an edge aggregation function $\rho_{E}^{(l)}$ and a node aggregation function $\rho_{V}^{(l)}$, as well as an edge combination function $\phi_{E}^{(l)}$ and a node combination function $\phi_{V}^{(l)}$. The combination functions may be neural networks involving matrices of weights $W^{(l)}_{E}$ and $W^{(l)}_{V}$ that are generally common to all edges and nodes, respectively. The outputs of an arbitrary intermediate layer $l$, given by its  combination function, are hidden feature vectors $h_{v}^{(l)}$ and $g_{e}^{(l)}$. At the end of the GNN, besides obtaining the output node and edge feature vectors, $h_{v}^{L}$ and $g_{e}^{L}$, there are global aggregation and combination functions $\rho_{G}$ and $\phi_{G}$, respectively, that provide final global output vector $\hat{y}$. \hl{Although most works assume that the graph is static, the computation may be repeated several times with evolving weight matrices to adapt to dynamic graphs} \cite{pareja2020evolvegcn}.

We finally note that, due to the emergence of GNNs, aggregation and combination functions have taken different names in the literature. In an attempt to unify the notation, some equivalences are listed in Table \ref{tab3}.\vspace{-0.1cm} 

\begin{table}[!htb]
    \centering
    \caption{Homogenized nomenclature for aggregate and combine functions in the literature}
    \vspace{-0.4cm}
    \begin{tabular}{|c|c|c|} \hline
         \textbf{Aggregation} & \textbf{Combination} & \textbf{Ref.}  \\ \hline
            Local transition & Local output & \cite{Scarselli2009} \\ \hline
            \multicolumn{2}{|c|}{Aggregators} & \cite{Hamilton2017a} \\ \hline
             Aggregation & Update & \cite{Battaglia2018} \\ \hline
             Message + Aggregate & Update & \cite{Liao2018} \\ \hline
             Message &  Update & \cite{Gilmer2017, fey2019fast} \\ \hline
             Message, reduce & Update & \cite{wang2019deep} \\ \hline
             Scatter + ApplyEdge + Gather & ApplyVertex & \cite{Ma2019} \\ \hline 
             Aggregation & Feature extraction + update & \cite{EnGN} \\ \hline
             Gather + Reduce & Transform + Activate & \cite{greta} \\ \hline
             Aggregation & \acrshort{dnn} computation & \cite{Auten2020} \\ \hline
             Aggregation & Embedding & \cite{AWB} \\ \hline
             Aggregation &  Combination & \cite{Yan2020,Yan2020a} \\ \hline
             
    \end{tabular}
    \vspace{-0.5cm}
    \label{tab3}
\end{table}

\subsection{General Structure}
\label{sec:structure}
\begin{figure*}
    \centering
    \includegraphics[width = \textwidth]{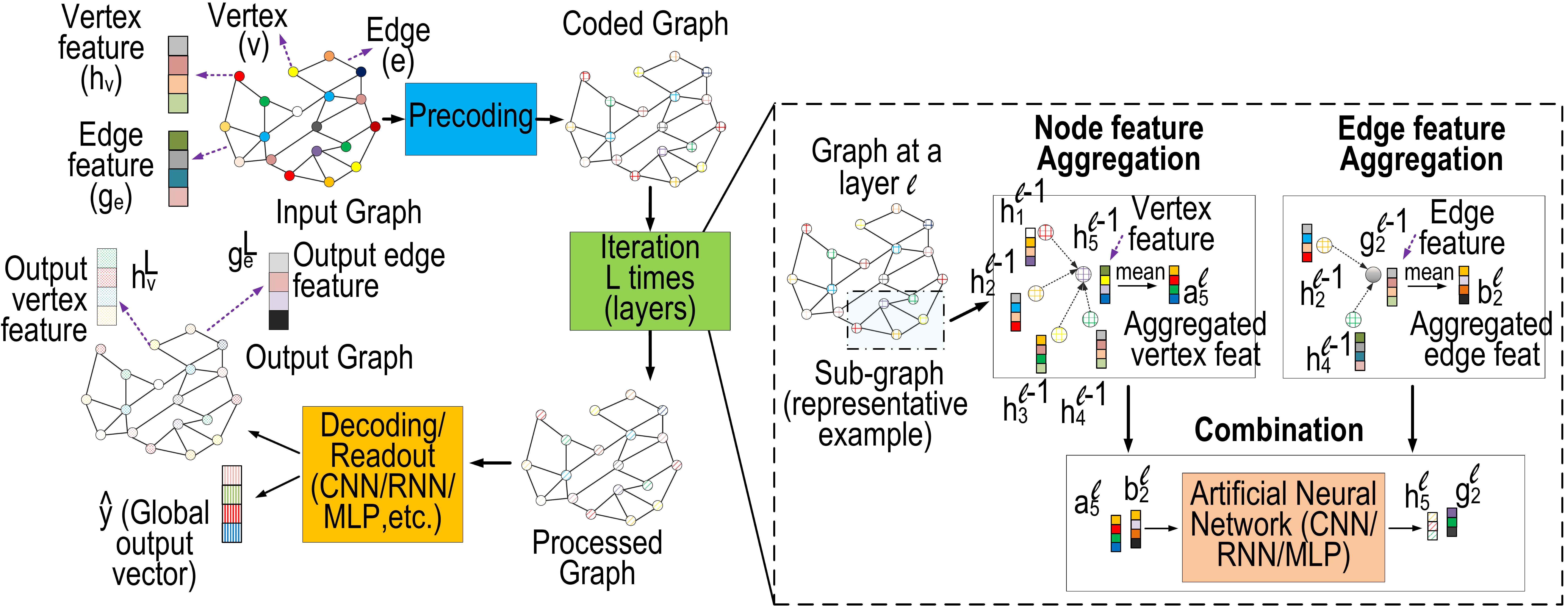}
    \vspace{-0.6cm}
    \caption{\acrshort{gnn} execution stages during inference: pre-coding, iterative process, and readout.}
    \vspace{-0.6cm}
    \label{Fig3}
\end{figure*}

Fundamentally, a GNN is an algorithm that leverages the graph connectivity to learn and model the relationships between nodes. Through an iterative process that depends on the graph structure, the GNN takes the input edge, vertex, and graph feature vectors (representing their known attributes) and transforms them into output feature vectors (representing the target predictions).  
In general, the GNN operation contains the steps illustrated in Fig. \ref{Fig3}:
\begin{enumerate}
    \item \textbf{Pre-processing:} this is an initial and optional step generally done offline that can transform the input feature vectors and graph structure representation through a precoding process. This may be used to sample the graph, \hl{to re-order the graph towards} reducing the algorithm complexity \hl{and its processing}, or to encode the feature vectors, among others \cite{Zhang2020a,Hamilton2017a,Jia2020,Tian2020,chiang2019cluster,chen2020rubik, zeng2019graphsaint}.
    \item \textbf{Iterative updates:} After the pre-processing, the feature vectors of each edge and vertex are updated via the aggregate--combine functions iteratively. To update the edges, attributes from the edge itself, the connected vertices, and the graph are \textit{aggregated} into a single set and \textit{combined} to yield a new edge feature vector. Similarly, updating the vertices implies \textit{aggregating} the feature vectors from neighboring vertices $N(v)$ and \textit{combining} them to obtain a new feature vector. Note that each step or \emph{layer} updates each edge and vertex with information coming from neighbours located at a single hop. Thus, the iterative process allows to gradually account for relations of increasingly distant nodes and edges. \hl{Additionally, in each successive layer, the graph may be coarsened by means of pooling} \cite{ying2018hierarchical} \hl{or the neighbourhood set changed by means of layer sampling} \cite{Hamilton2017a}.
    \item \textbf{Decoding or readout:} if the graph has a global feature vector, it is updated once after the edge and node updates are completed. The final output is either an edge/node embedding, which is a low dimensional feature vector that represents edge- or node-specific information, or a graph embedding summarizing the information about the entire output graph instead. 
\end{enumerate}


As in any other neural network, the GNN processing depends \hl{on} its architecture. GNNs are basically divided into \emph{layers}, with each layer corresponding to one of the iterations in the update process described above. This means
that each layer allows information from nodes to propagate further away from it. Hence, the precise number of required layers will depend on how relevant are the relations among \emph{distant} nodes in a given application. \hl{The most widespread GNN algorithms have 1--5 layers} \cite{Hamilton2017a,Kipf2019,Velickovic2018,Xu2019, Rahimi2018SemisupervisedUG} as an excessive amount of layers typically lead to \hl{the problems of feature oversmoothing, vanishing gradients, or overfitting} \cite{Li2018b}. \hl{A few works have proposed techniques to alleviate these issues and enable deep GNNs of up to 100 layers} \cite{li2019deepgcns, chen2020simple}\hl{, yet these proposals are in their infancy.}


In each of the layers, information flows between vertices using an \textit{aggregation} function and feature vectors are updated via the \textit{combination} function after aggregation \hl{in a process similar to that of the classic Weisfeiler-Lehman (WL) test for graph isomorphism} \cite{Weisfeiler1968}. \hl{The size of aggregation depends on the number of vertices and edges (ranging from hundreds to billions) whereas the size of combination depends on the length of the feature vectors (ranging from dozens of features to tens of thousands)}. The aggregation and combination functions for both edges and vertices \hl{are crucial design decisions as they determine the expressive power of the GNN, which has been demonstrated to be} \emph{at most} \hl{equal to the WL test in distinguishing graph structures} \cite{Xu2019}. As we will see in Section \ref{sec:algorithms}, Table \ref{table_algorithms}, \hl{there is a wide variety of such functions ranging from simple averaging to weighted sums with learnable attention coefficients, different types of neural networks, from MLPs to LSTMs with their own weighted sums and non-linear activations, whose suitability depends on the relation to be learnt.} The operations may vary across layers and differ between edges, vertices, or global updates. However, the structure is often simplified by (i) sharing the same operation across layers and (ii) removing or considering trivial combination functions for the updates of edges or nodes. 

\hl{The fundamental structure here explained and depicted in Figure} \ref{Fig3} \hl{can be complemented with sampling and pooling operations which help to reduce the computational complexity of GNNs} \cite{Hamilton2017a,ying2018hierarchical,zeng2019graphsaint}, \hl{and/or augmented with support for dynamic graphs} \cite{pareja2020evolvegcn}. \hl{Sampling refers to the pruning of either the graph or the neighbourhood set of each node, and it is used to limit or harmonize the resources and runtime of the aggregation process, whereas pooling refers to the coarsening of the graph from one layer to the next, thus reducing the amount of nodes to process in both aggregation and combination. To add support for dynamic graphs, whose structure and input feature vectors may evolve over time, recurrent units are generally used to adapt the weight matrices in each time step.}

In summary, we can understand \acrshort{gnnpl} as a collection of neural networks working over a graph's connectivity. In the scope of each layer, we have up to two neural networks with learnable weights that determine the combination of edges and vertices, respectively. In the scope of the whole GNN, we have a neural network with learnable weights that determines the global update. The way these operations take place for inference and training is depicted next. 


\subsection{Computing GNN Inference} \label{sec2b}
Algorithm \ref{alg1} shows a pseudo-code describing GNN inference. The algorithm may take as inputs the feature vectors of the edges, vertices, and graph; or initialize them. We can see how the execution is divided into layers (line 9) and, within each layer, each and every edge is updated in parallel by aggregating its own feature vector with those of the connected vertices (line 11). Each and every vertex is also updated in parallel by aggregating the feature vectors of its neighbours with itself (line 15). The aggregated edges and vertices are transformed via combination functions (lines 13 and 17), which can be neural networks as we see in Section \ref{sec:algorithms}. Following the completion of the iterative process, a readout is performed using the corresponding function, which may again possibly be a neural network (line 18). 

For an arbitrary layer $l \in [1,L]$, edge transformation occurs as
\begin{equation}
    \text{AGGREGATION:   }\quad b_e^{(l)} = \rho_{E}^{(l)}(\{g_e^{(l-1)},h_u^{(l-1)}: u\in N(e)\}),
    \end{equation}
    \begin{equation}
\text{COMBINATION:   }\quad g_e^{(l)} = \phi_{E}^{(l)}(\{b_e^{(l)}\}),
\end{equation}
so that the aggregation of edges $\rho_{E}$ takes the feature vector $g_{e}$ of the edge itself $e$, as well as the feature vectors of the vertices at its endpoints, $h_{u}$ with $u \in N(e)$, for the previous layer $l-1$. The combination $\phi_{E}$ uses this aggregation as input \cite{Xu2019}. A similar reasoning applies to the aggregation and combination of vertices
\begin{equation}
    \text{AGGREGATION:   }\quad a_v^{(l)} = \rho_{V}^{(l)}(\{h_v^{(l-1)},h_u^{(l-1)}:u\in N(v)\}),
    \end{equation}
    \begin{equation}
\text{COMBINATION:   }\quad h_v^{(l)} = \phi_{V}^{(l)}(\{a_v^{(l)}\}).
\end{equation}
The equations describe how $a_{v}^{(l)}$ is calculated as the aggregation of the feature vectors from the nodes that are neighbours to $v$, from the previous layer $l-1$, and how the feature vector of layer $l$ is calculated using the aggregation $a_{v}^{(l)}$ as input. Lastly, a final readout function is applied, which may involve the aggregation and combination of feature vectors from edges and vertices of the entire graph, and from the last iteration $L$, hence obtaining the output feature vector $\hat{y}$ as
\begin{equation}
\text{READOUT:   }\quad \hat{y} = \phi_{G}(\rho_{G}(\{h_{v}^{L}, g_{e}^L : v,e \in G\})).
\end{equation}

\begin{algorithm}
\caption{\acrshort{gnn} Operations in Inference \label{alg1}} 
\begin{algorithmic}[1]
\Procedure{GNNOperator}{}
\State $L \gets$ \text{Number of layers in the GNN}
\State $V \gets$ \text{Set of nodes in graph} $G$  \hl{(assumed static)}
\State $E \gets$ \text{Set of edges in graph} $G$ \hl{(assumed static)}
\Statex \textbf{Initialize Nodes and Edges:}
\For{$v \in V$}
\State $h_{v}^{0} \gets$ [$x_{v}$,0,$\ldots$,0]
\EndFor
\For{$e \in E$}
\State $g_{e}^{0} \gets$ [$z_{v}$,0,$\ldots$,0]
\EndFor
\Statex\textbf{GNN Layered processing:}
\For{$l = 1$ to $L$} 
\Statex\textbf{Edge processing:} \hl{// Order of edge and node processing may be interchanged or even interspersed.}
\For{$e \in E$} \hl{// Order of aggregation and combination may be interchanged if aggregation is linear.}
\State $ b_e^{(l)} = \rho_{E}^{(l)}(\{g_e^{(l-1)},h_u^{(l-1)}: u\in N(e)\})$
\State $g_e^{(l)} = \phi_{E}^{(l)}(\{b_e^{(l)}\})$
\EndFor
\Statex \textbf{Node processing:} \hl{// Order of edge and node processing may be interchanged or even interspersed.}
\For{$v \in V$} \hl{// Order of aggregation and combination may be interchanged if aggregation is linear.}
\State $a_v^{(l)} = \rho_{V}^{(l)}(\{h_v^{(l-1)},h_u^{(l-1)}:u\in N(v)\})$
\State $h_v^{(l)} = \phi_{V}^{(l)}(\{a_v^{(l)}\})$
\EndFor
\EndFor
\Statex \textbf{Readout:} 
\State $\hat{y} = \phi_{G}(\rho_{G}(\{h_{v}^{L}, g_{e}^L : v,e \in G\}))$
\EndProcedure
\end{algorithmic}
\end{algorithm}


Algorithm \ref{alg1} hinges on the general assumption that aggregation and combination functions are (i) invariant to permutation of nodes and edges, since there does not exist any implicit order in a graph structure, unless some node feature indicates such an order; and (ii) invariant to the number of input nodes, since the degree of nodes may vary widely across the graph \cite{Battaglia2018}. This implies that the functions \hl{within a layer can be applied to all edges and all vertices in parallel, following any order}. Further, the order between aggregation and combination can be switched if the aggregation function is linear \cite{EnGN}. However, \hl{it is important that the order of layers is preserved to avoid violating data dependencies, which implies that all edge and node operations of layer $l$ shall finish before starting those of $l+1$.}

\begin{figure*}
    \centering
    \includegraphics[width=\textwidth]{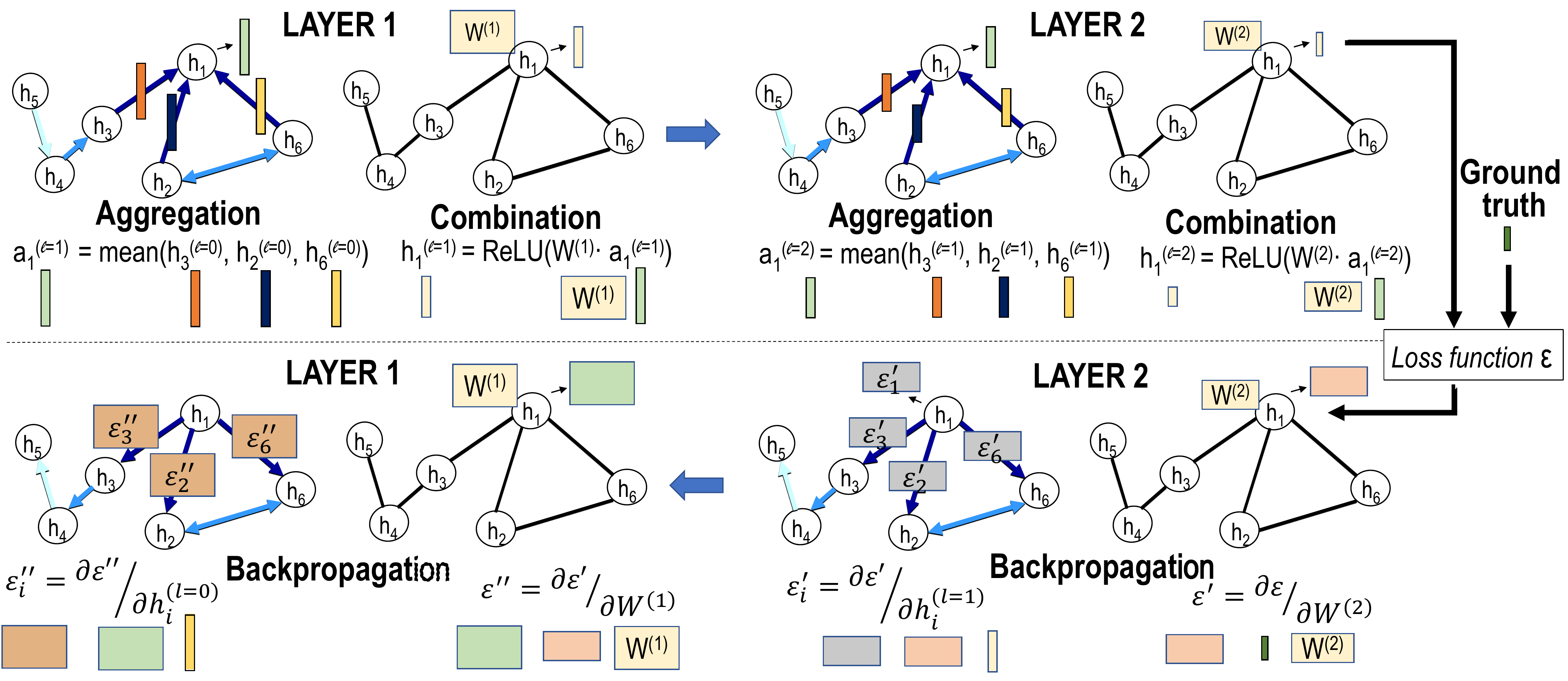}
    \vspace{-0.7cm}
    \caption{Example of computation in a sample GNN with node-level aggregation in inference (top left to top right) and training (bottom right to bottom left). The GNN has two layers, mean as the aggregration operator, and weighted ReLu for the combination. We show operations for node 1 only.\label{Figinference}\label{Figtraining}}
    \vspace{-0.2cm}
\end{figure*}

To exemplify the computation occurring in inference, top charts of Figure \ref{Figinference} represent the layers of a simple GNN with vertex aggregation and combination only. We show the operations from the perspective of node 1, although all nodes would be realizing the same computations concurrently. We illustrate how the graph connectivity drives the aggregation from nodes 2, 3, and 6 into node 1, and that combination reduces the length of the feature vector through the weight matrices $W^{(1)}$. \hl{We note, however, that combination functions do not necessarily reduce the length of the feature vectors; that depends on the actual GNN architecture.} The second layer repeats the exact same sequence, again reducing the length of the feature vector, this time through a different weight matrix $W^{(2)}$.

\textbf{Extended notation for sampling, pooling, and dynamic graphs:} \hl{As described above, sampling and pooling might impact the length aggregation and combination stages, whereas dealing with evolving graphs may require extra computation steps. Following the notation above, sampling essentially modifies either the input graph,} $G_{s}$ \cite{zeng2019graphsaint}\hl{, or the neighbourhood operator making it dependent on the layer being computed} $N_{s}^{(l)}(v)$. \hl{Pooling can be seen as a graph transformation across layers, thus making the set of nodes and edges to vary as well} $E^{(l)}$, $V^{(l)}$.
\hl{Finally, support for dynamic graphs} $G_t$ \hl{requires the entire GNN to be time-dependent, introducing time in the notation. Neighbourhood sets, feature vectors, and most importantly, weight matrices would evolve over time,} $N_t(\cdot)$,  $h^{(l)}_{v,t}$, $g^{(l)}_{e,t}$, $W^{(l)}_{E,t}$, $W^{(l)}_{V,t}$.  

\textbf{Message passing equivalence:} We note that notation relative to GNN algorithms is diverse in the literature. A notable example is that of \acrfull{mpnn} \cite{Battaglia2018}, which describes the aggregations as message passing functions $M(\cdot)$, the combinations as update functions $U(\cdot)$, or the layers as time steps. Table \ref{tabrelat} illustrates the equivalence between the MPNN formulations and the corresponding generic formulations from Eqs. (1)-(5).

\begin{table}
    \centering
    \caption{Equivalence between general and Message Passing Neural Network (MPNN) formulations}
    \vspace{-0.4cm}
    \setlength\tabcolsep{3pt}
    \begin{tabular}{|m{3.3cm}|m{3.8cm}|m{7cm}|}  \hline
         \textbf{General} & \textbf{MPNN} & \textbf{Comments} \\ \hline
         $l$ & $t$ & Layer, time step, or epoch \\ \hline
         $v$ & $v$ & Node or vertex of interest \\ \hline
         $u \in N(v)$ & $w \in N(v)$ & Node within the neighboring set $N(v)$ of node $v$ \\ \hline
         $h_u^{(l)}$ & $h_w^t$ & Feature vector of vertex $u$ at layer $l$ or epoch $t$ \\ \hline
        $\rho^{(l)}(\{h_u^{(l-1)}: u\in N(v)\})$ & $\sum_{w \in N(v)}$ $M_{t}(h_{v}^{t-1},h_{w}^{t-1},e_{vw})$ & Aggregation at a layer or epoch with $M_{t}(\cdot)$ and $\rho^{(l)}(\cdot)$ as aggregation functions\\ \hline
         $a_v^{(l)} $ & $ m_{v}^{t}$ & Aggregated feature vector\\ \hline
         $\phi^{(l)}(\{a_v^{(l)}\})$ & $h_{v}^{t+1} = U_{t}(h_{v}^{t},m_{v}^{t+1})$ & Combination with functions $U_{t}(\cdot)$ and $\phi^{(l)}(\cdot)$ in a given layer or epoch\\ \hline
    \end{tabular}
    \vspace{-0.5cm}
    \label{tabrelat}
\end{table}

\textbf{Matrix multiplication notation:} \hl{GNNs are typically expressed in matrix notation that helps understanding the underlying computation. An example for node classification with sum aggregation function is as follows. Let} $A$ be the normalized adjacency matrix of the input graph, $H^{(l)}$ the matrix of features for layer $l$, and $W^{(l)}=W_{V}^{(l)}$ \hl{the weight matrix for the vertex combination function. Then, the forward propagation to layer} $l+1$ can be expressed as
\begin{equation}
H^{(l+1)} = \sigma(A H^{(l)} W^{(l)}),    
\end{equation}
where $\sigma(\cdot)$ \hl{is the non-linear activation function, e.g. a ReLU. For more complex GNNs and aggregation-combination functions, the forward propagation equation may change.}



\subsection{Computing GNN Training}
Aggregation, combination, and readout functions can be neural networks that may need to be trained before deployment. Training is performed via modifications of the traditional backpropagation algorithms, which take into account the unique traits of a GNN. Since a GNN unfolds into $L$ layers similarly to a \acrshort{rnn}, most GNNs employ \acrfull{bptt} schemes or variants of it. A popular variant of \acrshort{bptt} is the Pineda-Almeida algorithm \cite{Pineda,Almeida1987ALR}, which relaxes the memory requirements as already mentioned in the seminal work by Scarselli \emph{et al.} \cite{Scarselli2009}.

Specifically, in \acrshort{bptt}, a forward pass is first performed on the unfolded version of the GNN with its $L$ layers. The loss function $\varepsilon$ is then computed and the necessary gradient is backpropagated across layers. Since the weights are shared among all $L$ layers, they are updated accordingly. This process is carried out recurrently with multiple samples, often grouped in batches, until some target accuracy is reached. \hl{Depending on the problem, a sample can refer to the entire graph (e.g. representing a specific molecule) or a portion of it (e.g. a set of users in a recommendation system).}

To exemplify the computation occurring during training, bottom charts of Figure \ref{Figinference} represent backpropagation in a two-layer GNN. Again, we show the operations from the perspective of node 1, although all nodes would be realizing similar computations at the same time. \hl{The backward pass implies calculating the gradient of the loss function} with respect to the weights first, via partial derivative over $W^{(2)}$, and then with respect to each vertex's feature vector. The operation is then repeated for the first layer, via its own weight matrix $W^{(1)}$ and each vertex's feature vector. The derivatives of the loss function are, eventually, used to update the weight matrices.

The computation of the loss function depends on the type of learning. 
While graph-centric approaches tend to be trained using supervised learning, node-centric approaches are usually trained by means of semi-supervised learning, wherein information of the node features from a portion of the graph, and not the whole graph, is utilized. An example of the former method can be learning if a specific new molecule (graph) has a certain property, using a \acrshort{gnn} trained with molecules (graphs) whose properties are known beforehand and used as ground truth \cite{Gilmer2017}. For the latter method, an example can be a recommender system. In such a system, a graph represents a store with nodes being shopping items and their features, and edges being relations among items. The output feature vector could describe how likely a given user will be satisfied with a particular item. In this case, a priori complete information is not available and semi-supervised learning from past purchases by this and other users (a subgraph) is used instead \cite{Ying2018}.

\textbf{Matrix multiplication notation:} \hl{To express backpropagation in a compact manner, we adapt the formulation of} \cite{Tripathy2020} \hl{to the notation introduced in the previous section. Let }$Z^{(l)} = A H^{(l)} W^{(l)}$ so that $H^{(l+1)} = \sigma(Z^{(l)})$. \hl{Then, the backpropagation starts by calculating the gradient of the loss function} $\varepsilon$\hl{, which we denote as} $Y$\hl{, with respect to the weight matrix of the last layer. For an arbitrary layer $l$, this operation yields}
\begin{equation}
Y^{(l-1)} = \frac{\partial \varepsilon}{\partial W^{(l)}} = (H^{(l-1)})^{T} A G^{(l)},
\end{equation}
where $G^{(l)}$ is the gradient with respect to $Z^{(l)}$ and $^{T}$ denotes a transpose matrix. Therefore, $G^{(l)}$ \hl{refers to the propagation of the error back to each particular aggregated feature vector, yielding}
\begin{equation}
G^{(l-1)} = A G^{(l)} (W^{(l)})^{T} \odot \sigma'(Z^{(l-1)}), \text{\,\,\,with\,\,\,} G^{L} = \frac{\partial \varepsilon}{\partial Z^{(L)}} = \frac{\partial \varepsilon}{\partial H^{(L)}} \sigma'(Z^{(L)}),
\end{equation}
where $\sigma'$ is the derivative of the non-linear activation function.


\section{The Evolution of the GNN Field}\label{sec3}
In this section, \hl{we aim to} demonstrate that GNN computing is in an early \hl{yet rising} stage \hl{as compared to the rest of GNN disciplines. We also observe that there is a wide variety of GNN algorithms that, as we will see in Section} \ref{sec4}\hl{, complicate the task of designing accelerators. To these ends,} we present the evolution of the body of knowledge in the area of \acrshort{gnnpl} \hl{from a general perspective in Section} \ref{sec:taxonomy} \hl{and from an algorithm perspective in Section} \ref{sec:algorithms}. 

The study uses a \acrshort{kg} approach that naturally exposes the confluence of multiple interrelated sub-fields in the GNN landscape. To generate the \acrshort{kg}, \hl{a repository of annotated papers has been created. The papers are classified by their year of publication and are manually given a single tag using the title and keywords as main reference.} Further, the references of each paper are extracted by means of the CERMINE library \cite{tkaczyk2015cermine}. The generated database is introduced into the Neo4j graph tool \cite{webber2012programmatic}, which allows to visualize the \acrshort{kg} \hl{with nodes and edges representing papers and their citation relations, respectively. To highlight the category and importance of papers, vertices are color-coded depending on the paper category and sized proportionally to the number of citations.}


\subsection{A General Perspective}
\label{sec:taxonomy}
Our first treatment of the GNN literature consists in classifying the papers by discipline. Concretely, we define the following taxonomy with topics ranging from formal mathematical aspects, to the algorithms, applications, and computing aspects: \textit{GNN modeling}, \textit{GNN applications}, \textit{GNN complexity}, \textit{GNN algorithms}, \textit{GNN accelerators}, \textit{GNN HW/SW requirements}, and \textit{GNN dataflow}. The description of each topic, together with a discussion of its first works and the list of its references is given in Table \ref{tab:tags_table}. We also show the percentage of papers that pertain to a given category. 


\begin{table*}[t]
    \centering
    \caption{The different categories for the classification of the state of the art in \acrshort{gnnpl}.}
    \setlength\tabcolsep{2pt}
    \small
    \vspace{-0.4cm}
    \begin{tabular}{|m{1.4cm}|m{4.7cm}|m{5.7cm}|m{1.5cm}|p{1cm}|}
    \hline
Tag Name & Meaning & Origins & References & Fraction \\ \hline

\acrshort{gnn}\quad\quad modeling & This category includes the papers that encompass the topics of design and mathematical formulation of \acrshort{gnnpl}. Other salient design formalisms related to \acrshort{gnnpl} have been also categorized in this tag. & \textbf{2005.} While the most important paper in \acrshort{gnn} modeling is from Scarselli \emph{et al.} in 2009, it extends a seminal work from 2005. It defines the mathematical foundation of these GNNs, and thus becomes a fundamental paper in this category. & \cite{Kipf2019,Hamilton2017a,Scarselli2009,Gori2005,Jiang2019,Onoro-Rubio2017,Niepert2016,Schlichtkrull2018,Garcia-Duran2017,Hamaguchi2017,Henaff2015, Dwivedi2017}& 13.19\% \\ \hline

\acrshort{gnn}\quad\quad  applications & Papers with this tag elaborate upon the various applications of \acrshort{gnnpl}, regardless of the field.
& \textbf{2005.} Given the ubiquity of graphs in real-world data, this is one of the first sub-fields to have emerged. In their seminal work, Scarselli \emph{et al.} presented the first possible applications together with the first GNN model \cite{Scarselli2005}. Since then, many other applications have appeared. & \cite{Ying2018,Duvenaud2015,Myska2019,Rusek2019,StJohn2019,Sanchez-Lengeling2019,Monfardini2006,Zhang2018,Scarselli2005,Fan2019,Verma2019,Li2017,Ma2019a,Zhu2018,Sanchez-Gonzalez2018,Battaglia2016,Cui2019,Fout2017,Zayats2018,Guo2019,Qiu2018,Kim2020}&24.17\%        \\ \hline

\acrshort{gnn} \quad\quad complexity & This tag encompasses the papers that explore the complexity within the \acrshort{gnn} structure and its operations. & \textbf{2009.} The exploration of the complexity of \acrshort{gnn} execution may have started with \cite{Scarselli2009a} in 2009, which analyzed the complexity for the most common \acrshort{gnn}s at that moment. After this, we have to wait until 2017 to find more works that take into account complexity, as datasets become more resource demanding and large-scale applications become apparent. & \cite{Scarselli2009a,Chen2018,Bianchini2005,Bruna2014,Wu2019a,Ying2019,Kawamoto2018,DiMassa2006,Zugner2019,Chen2019,Verma2019a, Nickel2015, Pineda}&14.29\%    \\ \hline

\acrshort{gnn} \quad\quad algorithms & This tag refers to papers that introduce new algorithm variants to the GNN family, including aspects such as attention, isomorphism, sampling, or new operations at the aggregate--combine phases. & \textbf{2009.} We consider \cite{Scarselli2009} as the first unification of multiple similar prior approaches. Others have attempted to do similar generalizations, such as the MPNN from Gilmer \emph{et al.} \cite{Gilmer2017} or the GN from Battaglia \emph{et al.} \cite{Battaglia2018}. & \cite{Yang2019,Hamilton2017,Chen2018a,wu2019comprehensive,Li2016,Kipf2016,Gilmer2017,Rusek2019a,Bandinelli2010,Zhang2019,Monti2017,Son,Battaglia2018,Ma2018a,Park2019,DeCao2018,Li2018,Thekumparampil2018,Garcia2018,Zhou2018a,Xu2019,Wang2018,Velickovic2018}&25.27\% \\ \hline

\acrshort{gnn}\quad\quad accelerators& Under this tag, we gather papers that target the acceleration of \acrshort{gnnpl} either via software or hardware. & \textbf{2017.} The earliest paper to tackle the problem of \acrshort{gnn} acceleration is \cite{Hamilton2017a}, in 2017, through a simplification of the algorithm via  sampling. More recent works on software in CPUs and GPUs, and hardware acceleration in custom architectures, have also been considered. & \cite{EnGN,AWB,Auten2020, Zhang2020acc, Kiningham2020, Zhu2018ali, Zeng2020, Yan2020, chen2020rubik, gnnadvisor}&10.99\%  \\ \hline

\acrshort{gnn} HW/SW \footnotesize requirements & This tag gathers works that, with the increasing popularity of \acrshort{gnnpl} as well as the complexity of the data-sets, analyzed the actual computational needs required to address these challenges. & \textbf{2018.} This specific sub-field started to gain traction in 2018, with the first work leading to \cite{Ma2019} where the hardware and software efficiencies in executing \acrshort{gnnpl} were studied. & \cite{Ma2019,Tripathy2020,Balog2019, Tian2020, Jia2020, Jia2019, Yan2020a, Zhang2020b} &8.79\% \\ \hline

\acrshort{gnn} dataflow & Dataflow refers to the movement of data within the processing engine, which becomes crucial for the design of custom accelerators. Hence, under this tag we categorize the papers that formally describe possible dataflow solutions. & \textbf{2018.} Two primary works, i.e., \cite{Ma2019}, which covers scalability in the training, and \cite{Liao2018} which covers efficiency for partitioning of the graph data, emerged. & \cite{greta,Ma2019,Liao2018}&3.30\% \\ \hline
           
    \end{tabular}
    
    \label{tab:tags_table}
\end{table*}



An important finding from our analysis is that the percentage of papers being categorized for \textit{\acrshort{gnn} accelerators}, \textit{\acrshort{gnn} HW/SW requirements}, and \textit{\acrshort{gnn} dataflow} are\hl{ 10.99\%, 8.79\% and 3.30\%,} respectively. These categories mostly relate to the computing side of GNNs as they concern the analysis of computational requirements of GNNs, optimization of GNNs via software, and development of hardware accelerators. 
We thus observe that a very small percentage of the existing research has approached \acrshort{gnnpl} from the perspective of computing. We further note that the first works to deal with these topics date back from 2017, when the very first specific paper on \acrshort{gnn} acceleration was published. It can be therefore concluded that GNN processing is in its nascent stages of development. This is the main reason for computing aspects not being analyzed in depth in recent GNN surveys \cite{Battaglia2018,Zhou2018a,wu2019comprehensive,chami2020machine,Lamb2020,Bronstein2017,Hamilton2017,Zhang2020a}, which we aim to address in this work, and also represents an opportunity to make an early impact in the GNN research field.

\begin{figure*}
    \centering
    \includegraphics[width = 
    0.85\textwidth]{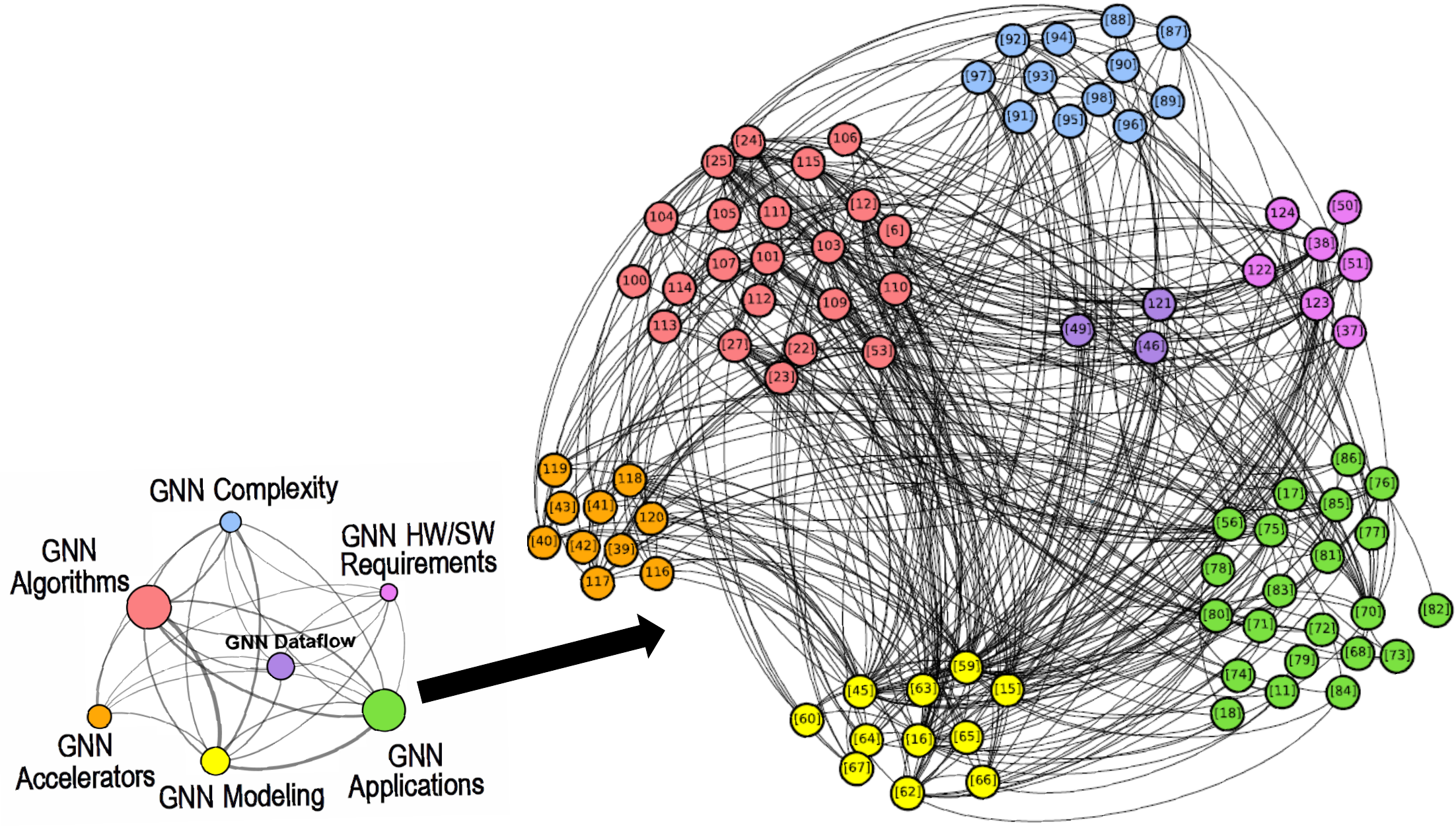}
    \vspace{-0.4cm}
    \caption{Full knowledge graph representation as of October 2020.}
    \vspace{-0.4cm}
    \label{fig:full_graph}
\end{figure*}

A second order analysis stems from the careful observation of the \acrshort{kg}, which we show in Fig. \ref{fig:full_graph}. 
\hl{In the left plot,} the size of the node represents the \hl{aggregated} number of papers in a category, whereas the thickness of the edge between two nodes illustrates the relative amount of citations between the papers of a given pair of categories. \hl{In the right plot,} we can also analyze the connections between the papers within the same category. Several observations can be made:
\begin{enumerate}[label=(\roman*)]
\item The categories related to computing are small yet well-connected to the theoretical side of GNNs, corroborating our earlier observation from Table \ref{tab:tags_table}.
\item The algorithms sub-field is large as many papers have appeared implementing multiple variants in the heterogeneous group of methods that GNN is. We review the evolution of GNN \hl{algorithms} later in Section \ref{sec:algorithms}.
\item The applications sub-field is large but sparsely connected internally, which means that application papers are generally not aware of other applications, unless reusing some specific common mechanism. This may be due to the wide variety of application fields for GNNs, ranging from social networks to chemistry, computer networks, or even material science as analyzed in previous sections.
\item The algorithm and application categories have a strong inter-connectivity, as each application paper shall at least mention the algorithms used to implement the proposed system.
\item The connection from application papers to computing papers is weak. This may be due to the relative immaturity of the GNN computing field and this may change in upcoming years, especially if applications clearly benefiting from specialized accelerators arise (akin to the appearance of CNN accelerators for computer vision).
\end{enumerate}


\begin{figure*}[t]
    \centering
    \includegraphics[width = 3.7cm]{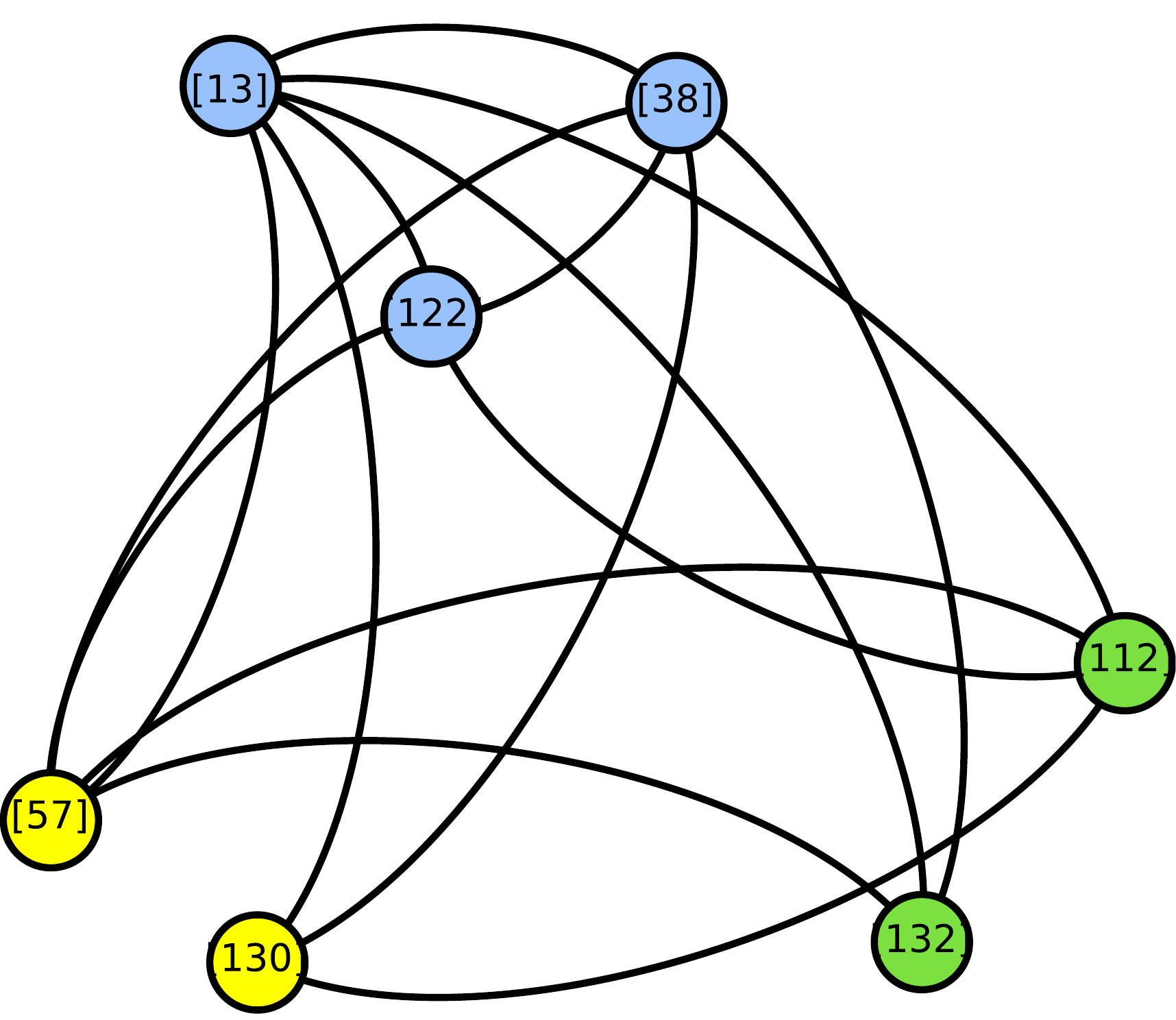}
    \includegraphics[width = 3.7cm]{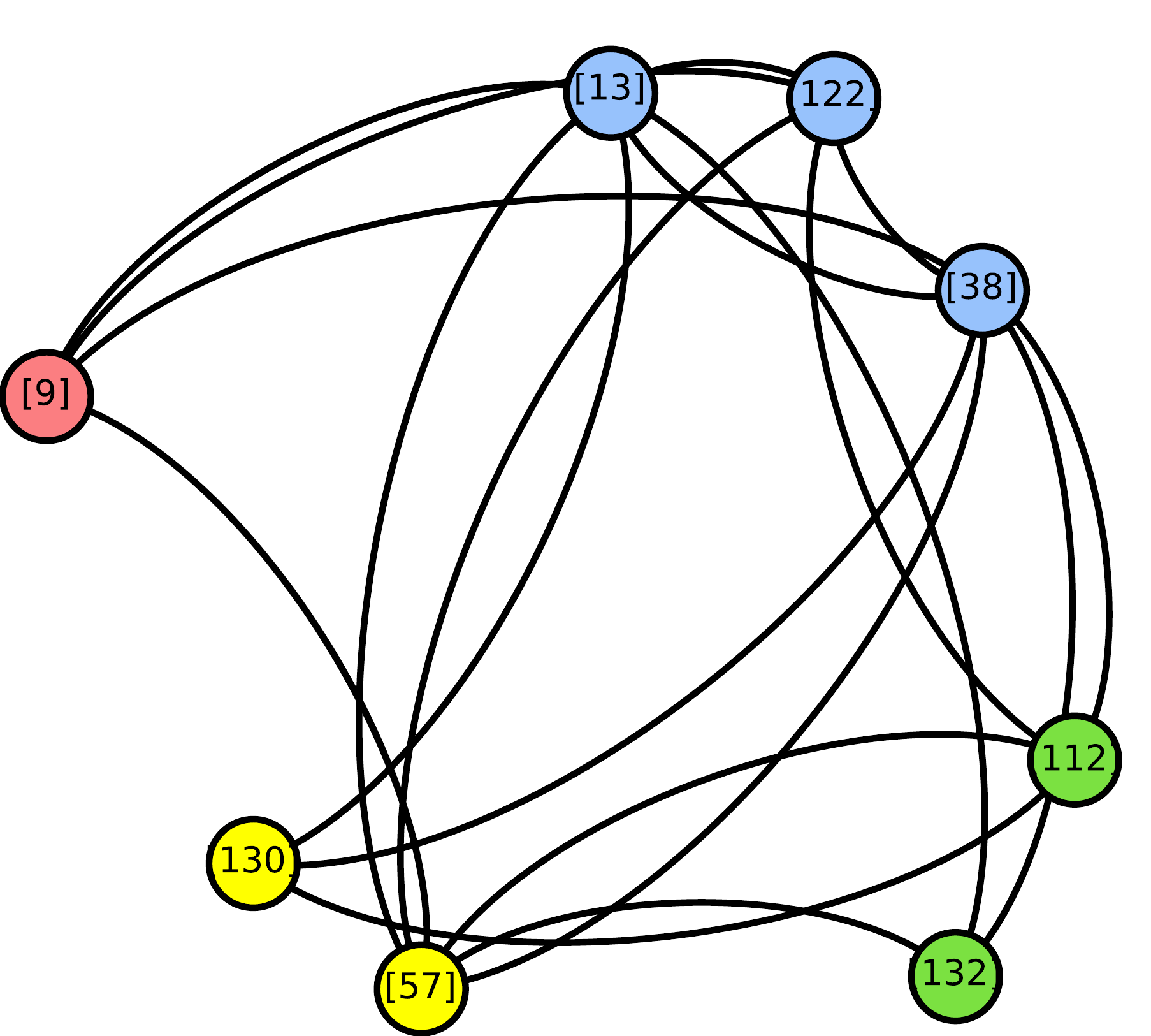}
    \includegraphics[width = 3.7cm]{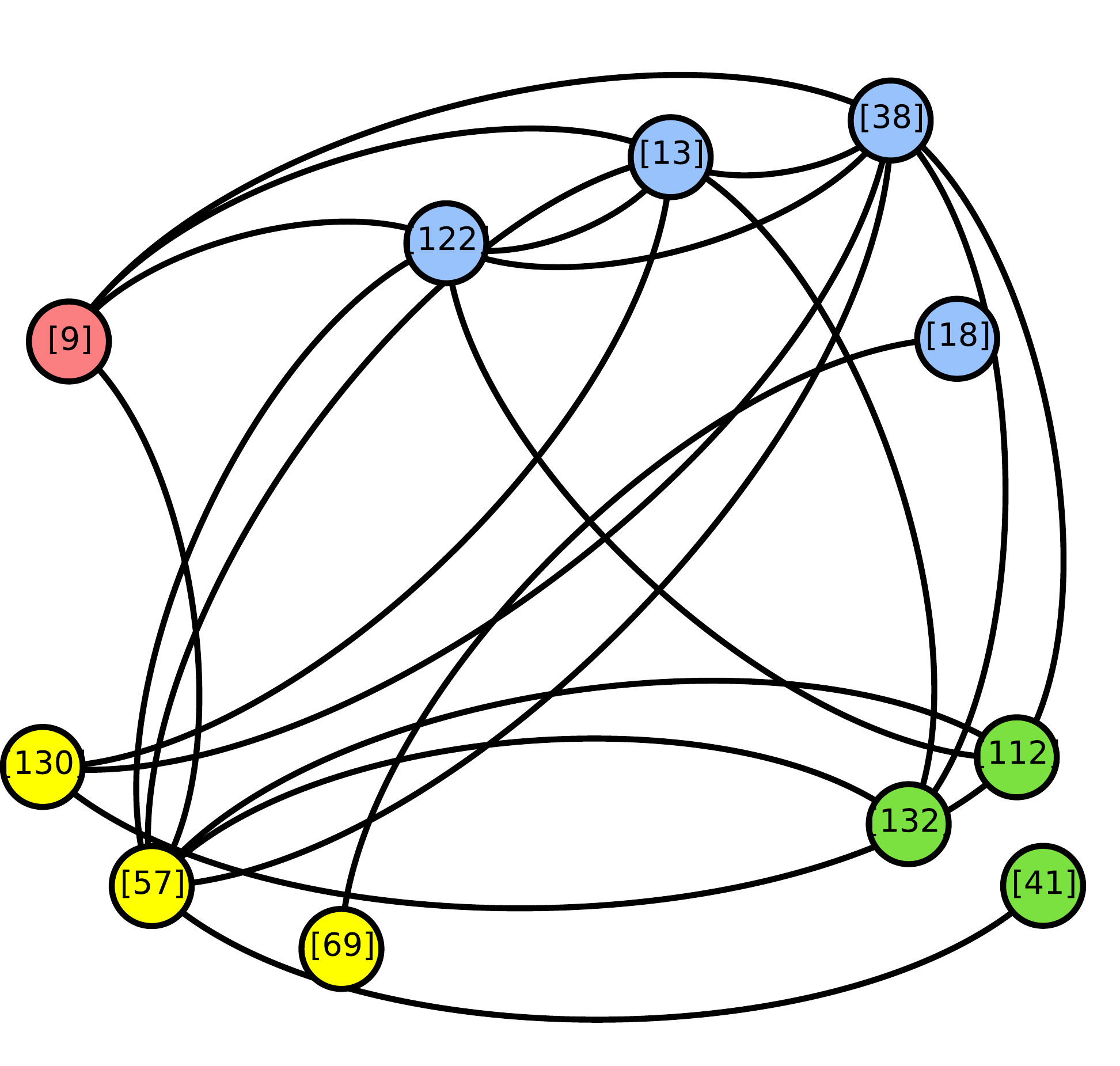}
    \includegraphics[width = 3.7cm]{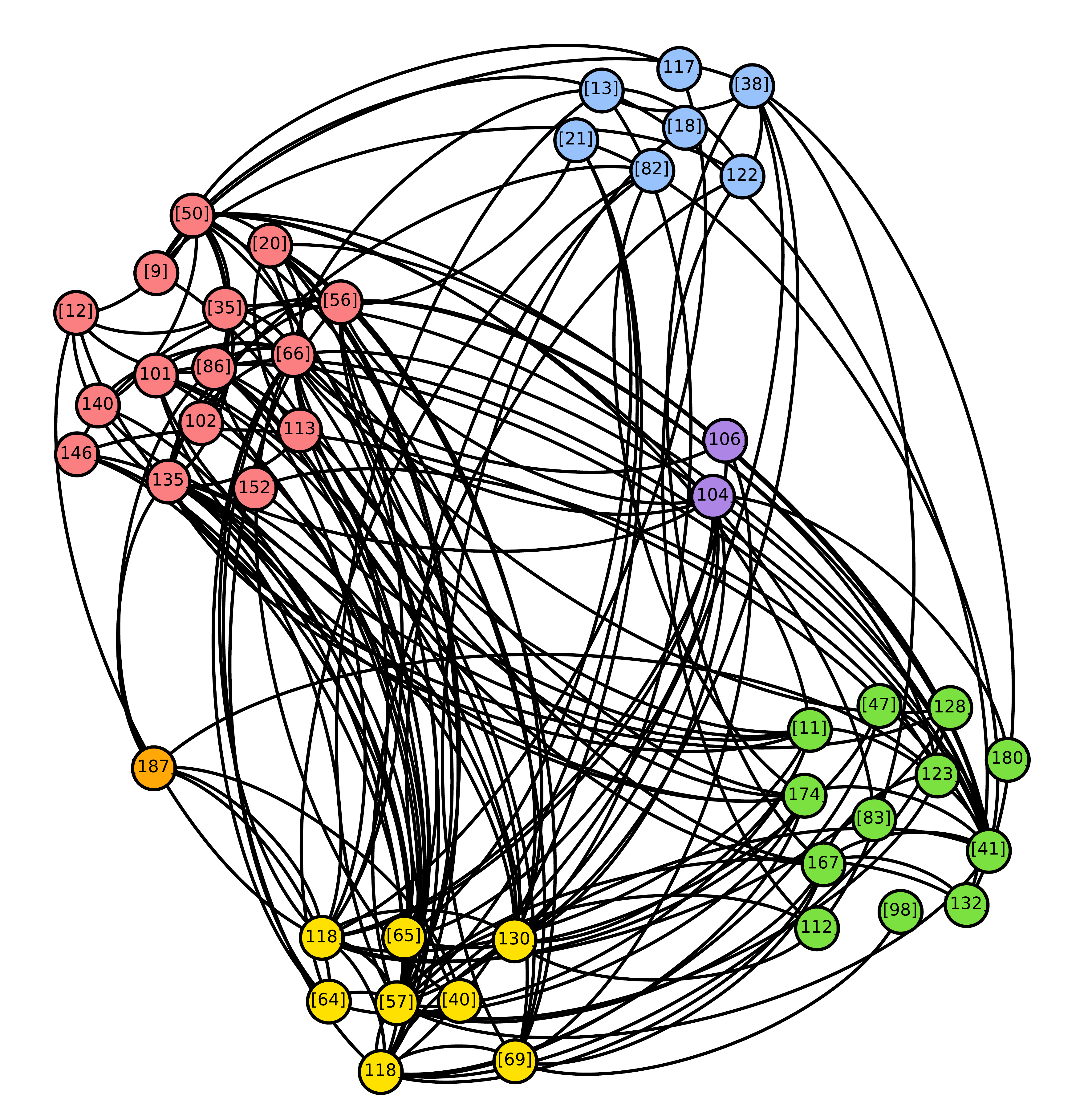}
    \includegraphics[height=4.8cm]{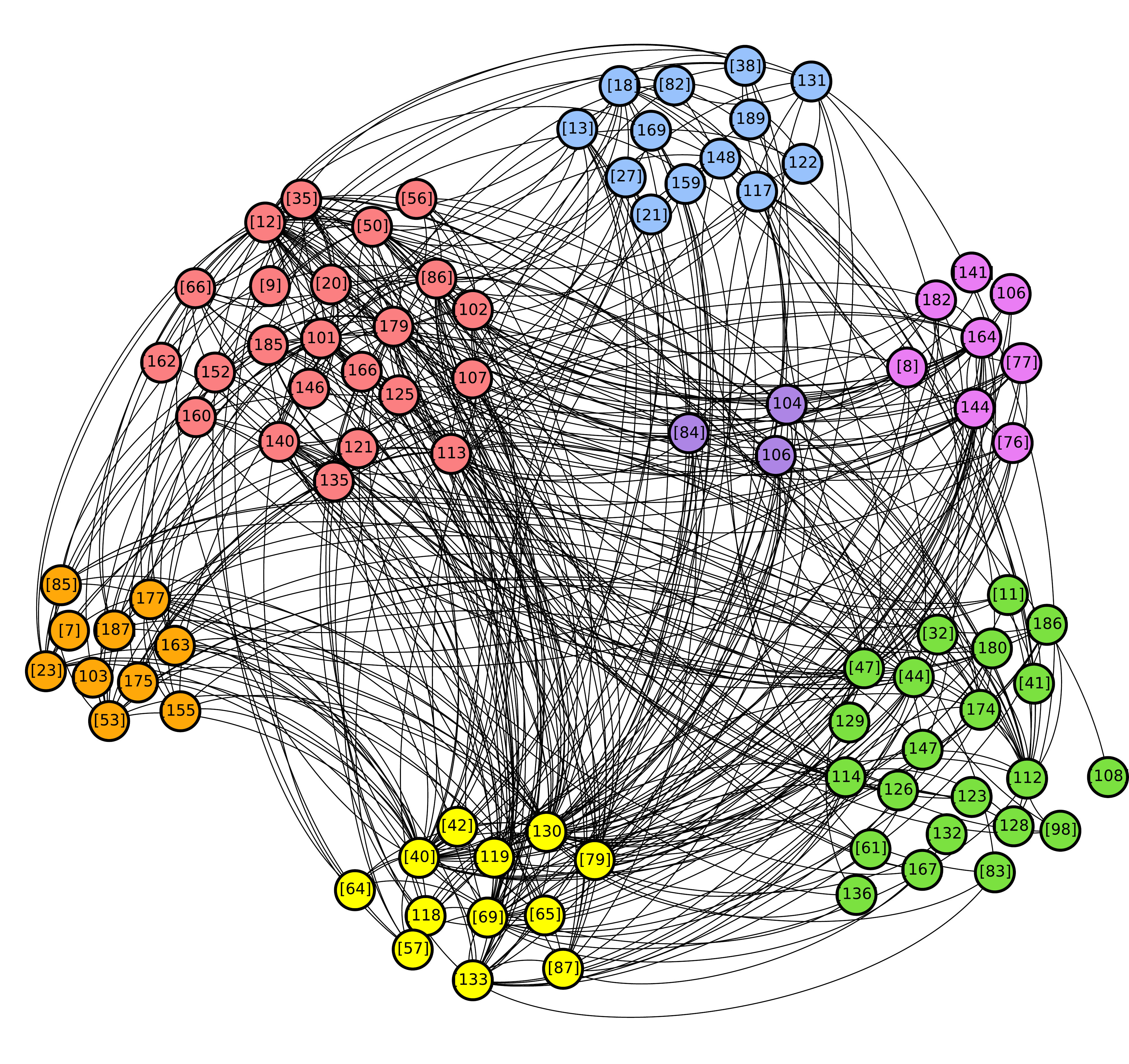}
    \hspace{0.5cm}
       \includegraphics[width=0.5\columnwidth,height=4.6cm]{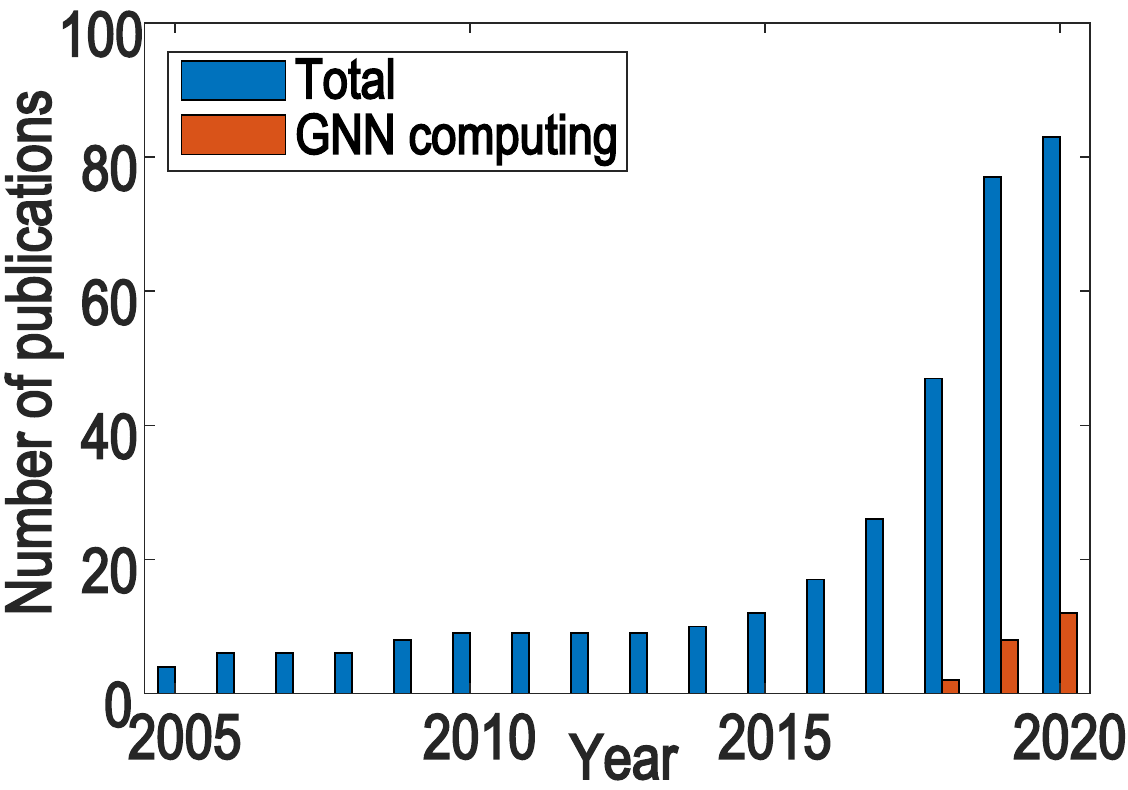}
    \vspace{-0.2cm}
    \caption{Evolution of the GNN knowledge graph over the years 2009, 2012, 2015, 2018, and 2020 (with the color code from Figure \ref{fig:full_graph}) and cumulative number of papers published in GNN in general and computing in particular.}
    \label{fig:graph_evolution}\label{fig:number_of_papers_evolution}
     \vspace{-0.2cm}
\end{figure*}

To further understand the state of things in \acrshort{gnnpl}, we visualize the evolution of the field over time. Specifically, we plot the growth of the \acrshort{kg} and of the amount of published papers over the years in Fig. \ref{fig:number_of_papers_evolution}. First works started to appear as soon as 2005 \cite{Gori2005} and, at that point, most research efforts were centered around new algorithms and possible applications. Evolution was rather slow for a decade, which we attribute to the lack of a killer application and the modest popularity of deep learning methods at that time. The field exploded around 2016, when CNNs and RNNs were already well established. Such a dramatic growth coincides with the introduction of the \acrfull{gcn} \cite{Kipf2016}, one of the first and most popular models for \acrshort{gnnpl}, later followed by the introduction of the message passing notation and quantum chemistry application in \cite{Gilmer2017}. 
We further observe that research on GNN computing started in 2017 and, since then, attained a similar growth to that of the field. This trend may be an indicator \hl{of} a strong increase of related works in the near future. Hence, it can be concluded that the area of \acrshort{gnn} accelerator design and development is emerging and, thus, necessitates deeper insights that we provide in upcoming sections.



\subsection{An Algorithm Perspective}
\label{sec:algorithms}
GNNs are a set of models with a vast amount of possible configurations and design decisions that allow to modulate the inductive bias of the algorithm. \hl{We have seen how,} due to their flexibility and potential applicability, \hl{the family of GNN algorithms has grown rapidly in recent years. Since different algorithms may be more or less amenable to certain acceleration techniques,} here we briefly summarize the progress in this sub-field \hl{from graph kernels to modern GNN algorithms}. Note that a deep review of existing GNN algorithms is not the main focus of this work. For such an analysis, we refer the reader to more specific surveys \cite{Battaglia2018,Zhou2018a,wu2019comprehensive,Zhang2020a,chami2020machine}.

\vspace{0.1cm} \noindent 
\textbf{Pre-GNN techniques.} 
Prior to the advent of \acrshort{gnnpl}, relational information extraction \hl{from graphs was based on graph embeddings, i.e. the pre-processing of the graph to condense the information in a low-dimensional space thus making it amenable to traditional ML algorithms} \cite{Bronstein2017, Cui2019b}. 
Similarly, \acrfull{gk} are a family of methods that, after extracting graph-level embeddings of two or more graphs, compare them for classification tasks \cite{Thomas2004, Ghosh2018}. 
\hl{An example of such approach is the random walk kernel,} wherein random walks are performed on the graphs while simultaneously counting the matching walks \cite{Tms}. 
As compared to \acrshort{gnnpl}, \acrshort{gk}s are easier to train because they have less hyperparameters, which on the other hand limits their performance. The main reason stems in the loss of potential information incurred by the process of graph embedding. Thus, to achieve acceptable performance, \acrshort{gk}s require handcrafted (not learned) feature maps, whilst \acrshort{gnnpl} do not. \acrshort{gnnpl} retain the inherent graph structure as a powerful and expressive form of defining the neural network, instead of distilling the essence of the graph to feed a conventional neural network.

\begin{figure} [!htb]
    \centering
    {\includegraphics[width=0.75\columnwidth]{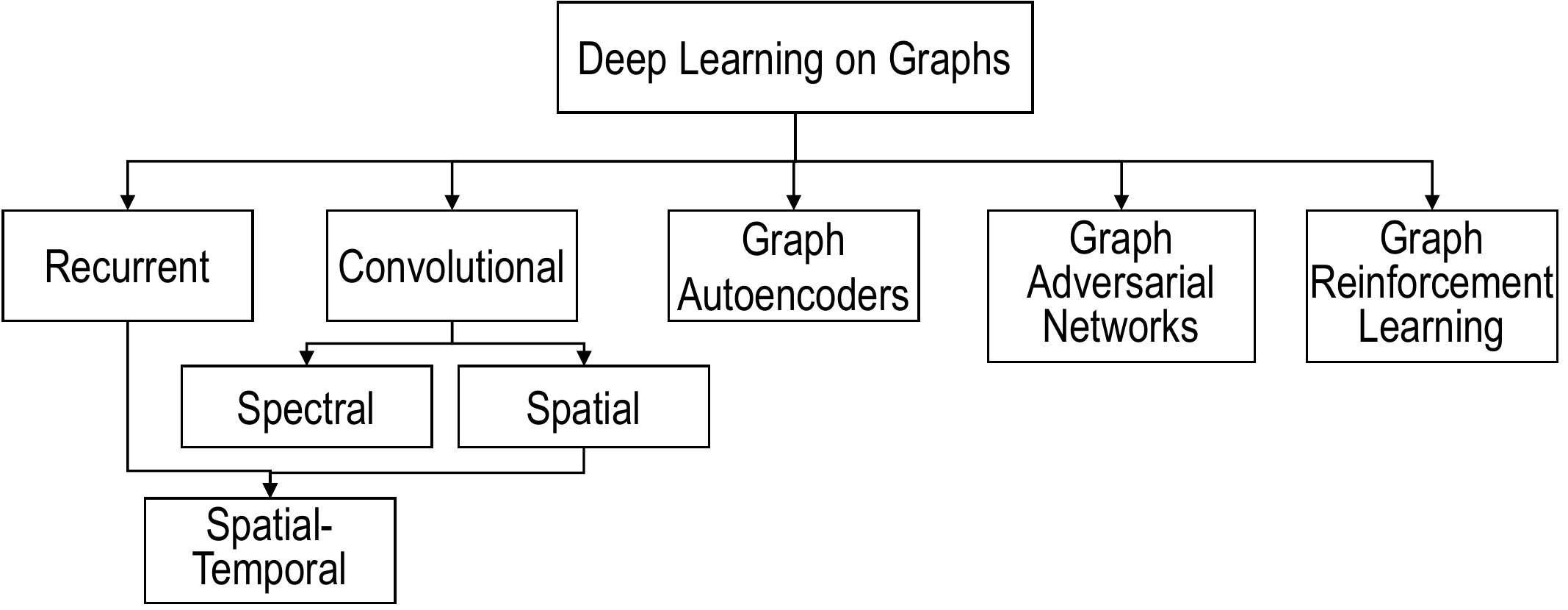}}
     \vspace{-0.2cm}
    \caption{GNN algorithm taxonomy based on model architectures and training strategies, adapted from \cite{Zhang2020a} and \cite{wu2019comprehensive}.}
    \label{Figflow}
     \vspace{-0.4cm}
\end{figure}

\vspace{0.1cm} \noindent 
\textbf{GNN algorithm classifications.} Since the seminal work by Scarselli et al. \cite{Scarselli2009}, multiple approaches have been published with the aim of elaborating and complementing the \acrshort{gnn} concept \cite{Niepert2016, Atwood2016, Bresson2016, Henaff2015} and many classifications can be carried out. 
A common distinction relates to the fundamental model upon which the GNN is built, \hl{for which a few taxonomies can be found in existing surveys} \cite{Battaglia2018,Zhou2018a,wu2019comprehensive,Zhang2020a,chami2020machine}. \hl{As a reference,} Fig. \ref{Figflow} reproduces the classification made in \cite{Zhou2018a} which mostly differentiates between recurrent-based \acrshort{gnnpl}, convolutional-based \acrshort{gnnpl}, \hl{graph autoencoders, graph reinforcement learning, and graph adversarial networks. We added the remark made in} \cite{wu2019comprehensive}, \hl{where combinations of recurrent and convolutional approaches are termed as spatial-temporal.}
 
On the one hand, recurrent-based \acrshort{gnnpl} refer to the initial \acrshort{gnn} models including that of Scarselli \cite{Scarselli2009}, which employ recurrent units as the combination function. Other examples are CommNet \cite{sukhbaatar2016learning}, which operates over simple aggregations without edge transformations, or Gated Graph Neural Networks (GG-NN, \cite{Li2016}), which use gated recurrent units \cite{Cho2014} as the update function to improve convergence. On the other hand, convolutional-based \acrshort{gnnpl} expand the idea of convolution in the graph space \cite{Chen} and can be divided into spectral-based \cite{Henaff2015} and spatial-based \acrshort{gnnpl} \cite{Zhu2018}. On the one hand, spectral-based models are built on spectral graph theory using graph signal processing techniques such as eigenvalue decomposition and filtering. However, they are computationally expensive methods, since the entire graph must be considered at once. On the other hand, spatial-based \acrshort{gnnpl} are much more computationally affordable, flexible, and scalable, since they only need to perform convolutions to the aggregation of features from neighbouring vertices \cite{Zhu2018}. 
Finally, spatial-temporal \acrshort{gnnpl} use both the spatial approach of the convolutions with the temporal approach of the recurrent units. An example is the network in gated graph convolutional network (G-GCN) from \cite{Bresson2018}.

\hl{Due to their flexibility and scalability, spatial-based convolutional} \acrshort{gnnpl} \hl{are arguably the most popular model} \cite{Chiang, wu2019simplifying, hu2019hierarchical, Schlichtkrull2018, Yu_2018, skeleton, Adaptive, Chen2018a, Gao_2018, tran2018filter}. In this paradigm, basic algorithms use a mean function as aggregation, sometimes also taking the degree of neighboring into account  \cite{Kipf2019}, after which many variants followed. GraphSAGE incorporated information of self-node features from previous layers in the update function and also pioneered the concept of sampling in GNNs to reduce the computational cost of aggregation \cite{Hamilton2017a}. FastGCN \cite{Chen2018a} also uses the sampling idea and integrates other strategies to speed up computations, such as evaluating integral formulations using Monte Carlo sampling. 
Another simplifying operation is the differential pooling of DiffPool \cite{ying2018hierarchical}, which forms hierarchical clusters so that later layers operate on coarser graphs. On a different approach, \acrfull{gin} \cite{Xu2019,Wu2019a} \hl{proved that the conditions needed for a GNN to achieve the maximum expressive power in capturing the structure of a graph are to emulate a WL test} \cite{Weisfeiler1968}\hl{. The particularity occurs at the graph output feature vector, which is obtained by concatenating the readout vectors of all layers.} We finally highlight \acrfull{gat} as the enabler of multiple works in \acrfull{nlp} \cite{Vaswani2017} and a particular case of the popular transformers approach. \acrshort{gat}s update the node features through a pairwise function between the nodes with learnable weights \cite{Velickovic2018}. This allows to operate with a learnt attention mechanism that describes the utility of the edges. 

Another branch of \acrshort{gnnpl} are the so-called \acrfull{gae} \cite{Kipf2016}. These \acrshort{gnnpl} are generative, which means that they convert a graph into a latent representation (i.e. encoding) that can be later expanded to generate to a new graph close in structure to the original one (i.e. decoding). What make these techniques unique in the graph domain is that \acrshort{gcn}s may be used to generate the low-dimensional vectors in the encoding process \cite{Salha2019}. 
\acrshort{gae}s are also typically trained using adversarial techniques, giving rise to graph adversarial networks such as NetRA \cite{yu2018learning}. 

We finally highlight that GNNs can be combined with reinforcement learning to give rise to novel graph learning techniques. For instance, MolGAN \cite{DeCao2018} generates molecular graphs with a certain end goal (reward). Another example is MINERVA, where reinforcement learning helps to predict the next node in the reasoning path of a KG \cite{das2018go}.


\vspace{0.1cm} \noindent 
\textbf{Comprehensive frameworks.} An aspect worth mentioning is that, within this multitude of algorithms, several groups have attempted to unify methods. One of the most popular ones is the message passing scheme \cite{Gilmer2017, zhang2020simple}, whose operation and description are amenable to convolutional networks for learning molecular fingerprints \cite{Duvenaud2015}, the classification methodology with \acrshort{gcn} from \cite{Kipf2019}, the interactive networks utilized for learning relationships and features \cite{Battaglia2016}, or also different flavours of Gated \acrshort{gnnpl}, to name a few. A further approach is that of the Non-Local Neural Networks (NLNN) \cite{Wang2018} aimed at unifying various attention approaches including GATs. These generally do not include edges features or aggregations and, instead, just involve pairwise scalar attention weights between nodes.  
Both MPNN and NLNN are also included into a further approach to unification referred to as Graph Networks (GNs) and proposed in \cite{Battaglia2018}. There, update functions applied to nodes, edges, or the complete graph are treated as differentiated blocks. The combination or repetition of several of these blocks gives rise to the different types of GNN found in the literature. \hl{Finally, Chami} \emph{et al.} \hl{propose an encoder-decoder model to express different graph embedding, graph regularization, graph auto-encoder, and GNN techniques} \cite{chami2020machine}.

\vspace{0.1cm} \noindent 
\textbf{Programming models.} From the perspective of computation, several programming abstractions are considered to support all possible operations within any GNN, generally compatible with the aggregate-combine model. \hl{These models are useful when the matrix multiplication notation cannot be employed because the aggregation or combination operations are not amenable to it, or because the adjacency matrix is extremely sparse and suggests the use of other representations such as compressed sparse row or column. In fact, as we will see in the next section, multiple accelerators implement GNN-oriented programming models.}

Among the different possible models, we highlight the \emph{Scatter-ApplyEdge-Gather-ApplyVertex with Neural Networks} (SAGA-NN) from \cite{Ma2019} \hl{which is followed implicitly in most modern libraries} \cite{Zhang2020b}. \hl{SAGA-NN augments classical scatter-gather approaches with two operations and works as follows: \texttt{Scatter} sends the nodes' feature vectors through their edges and \texttt{ApplyEdge} performs edge combination with the scattered vectors. Then, \texttt{Gather} allows each vertex to aggregate the vectors from its neighbours, and \texttt{ApplyVertex} performs the vertex combination after the gather operation. Another proposed model is that of} \emph{Gather-Reduce-Transform-Activate} (GReTA) from \cite{greta}. \hl{In this case, the four operations are user-defined and can be modified to implement any GNN. Aggregation is performed through \texttt{gather} and \texttt{reduce}, which allow each vertex to obtain the features from their neighbours and accumulate them into a single value. Combination is then performed through \texttt{transform} and \texttt{activate}, which typically do the matrix multiplication and non-linear activation of the aggregated data. More recently, Wang} \emph{et al.} proposed the \emph{NeighborSelection-Aggregation-Update} \hl{model, which adds a flexible neighbor selection layer to the more conventional aggregate-update} \cite{wang2021flexgraph}.

\section{The Revolution of GNN Acceleration}\label{sec4}
The optimization of ML algorithms and the building of custom hardware for high performance and efficiency has experienced an explosive growth in recent years \cite{resnet, eyeriss2}. This has come shortly after academia and industry have unveiled the outstanding potential of DNN algorithms and their all-pervasive applicability. As evidenced in previous sections, the field of GNNs is arriving at a similar turning point. At the time of this writing, research in GNN methods is already extensive and keeps refining the algorithms and investigating new applications with high potential impact. Therefore, a key research aspect in the years ahead will be how to compute GNNs efficiently to realize their full potential.  





\begin{table}
    \centering
    \caption{Operations in popular GNN algorithms.}
    \vspace{-0.4cm}
    \setlength\tabcolsep{2pt}
    \begin{tabular}{|m{3cm}|m{3cm}|m{3.2cm}|} \hline
         \textbf{Algorithm} &
         \textbf{Aggregation ($a$)} & \textbf{Combination ($h^{l+1})$} \\ \hline
             GCN \cite{Kipf2019} & $mean(N(h^l))$ &	$ReLU(W_l\cdot a)$ \\ \hline
             GIN \cite{Xu2019} & $mean(N(h^l))$ & $MLP(W\cdot ((1+\epsilon^l)\cdot h^l + a)$\\ \hline
             GS-mean \cite{Hamilton2017a} & $mean(N(h^l))$ & $\sigma(W_l\cdot Concat(a,h^l))$\\ \hline
             GS-max \cite{Hamilton2017a} & $max_{j\in N(h^l)}\sigma(W_l^1\cdot h^l_j)$ & $\sigma(W_l^2\cdot Concat(a,h^l))$\\ \hline
             GS-LSTM \cite{Hamilton2017a} & $LSTM(rand(N(h^l)))$ & $\sigma(W_l\cdot Concat(a,h^l))$\\ \hline
             GAT \cite{Velickovic2018} & $\sum_{j\in N(h^l)}\alpha_j W h_j^l $ &	$\sigma(a)$ \\ \hline
             HighwayGCN \cite{Rahimi2018SemisupervisedUG} & $\sigma(W^l\cdot h^l + b^l)$& $h^{l+1}\odot a + h^l\odot (1-a)$\\ \hline
             GRN \cite{EnGN}& $mean(N(h^l))$ & $GRU(h^l,W^l\cdot a)$\\ \hline
             
    \end{tabular}
    \\ \vspace{0.1cm} 
    Notation: $\sigma$ is a nonlinear function, $\alpha_j$ is the attention coefficient, $b$ is the bias, $\odot$ is a dot-product, $Concat$ is matrix concatenation, $MLP$ is a multi-layer perceptron, $GRU$ a gated recurrent unit,  and $LSTM$ is Long short-term memory.
    \label{table_algorithms}
    \vspace{-0.4cm}
\end{table}

GNN computing presents a set of unique challenges \cite{Yan2020a, Zhang2020b} that have rendered existing libraries and hardware platforms inefficient, including:

\begin{enumerate}[label=(\roman*)]
\item \textbf{The existence of multiple GNN variants, which may include edge, vertex, and graph-wide updates,} with a variety of aggregation and combination functions as illustrated in Table \ref{table_algorithms}, \hl{and possibly incorporating pooling and graph/layer sampling  operations as well} \cite{zeng2019graphsaint,chiang2019cluster}. These functions affect aspects such as the choice of operations to accelerate, the relative computational complexity of aggregation and combination, or the ordering constraints among them and across layers. \hl{Hence, instead of using a single general acceleration technique, GNN may require finding the right combination of techniques that works for a particular GNN variant.}

\item \textbf{The dependence of computation on the characteristics of the input graph} in terms \hl{of} size, sparsity, clustering, or the length of the associated feature vectors. Graph connectivity may follow a power-law distribution, be evenly distributed, or be bipartite. \hl{Since the computation fundamentally depends on the input graph,} decisions such as the use of dense or sparse logic, the dataflow to implement, the partitioning strategy, or the partitions' mapping and scheduling may need to be changed within and across graphs to maximize performance \cite{Tian2020, Jia2020, garg2021taxonomy}. The challenge is, therefore, to develop accelerators that can dynamically adapt to the graph characteristics.  
\item \textbf{A unique combination of computing characteristics from deep learning and graph processing, leading to alternate execution patterns.} \hl{More specifically, combination often implies MLP-like operations over a dense weight matrix, which is generally} computation-bound \cite{Sze2017}. \hl{In contrast, aggregation involves, among other operations, fetching groups of vertices that often lead to irregular memory patterns} \cite{Gui2019}. \hl{Optimizations in aggregation can be done via sparse GEMM of the adjacency matrix} \cite{Yan2020a}\hl{, but they are not generalizable to all graphs/GNNs and typically not enough to combat the extreme sparsity of adjacency matrices}. \hl{Therefore, the challenge is to develop architectures that accelerate such distinct phases and their intertwining at runtime.}

\item \textbf{A wide pool of applications with not only different graph characteristics, but also different performance targets.} \hl{For example, recommendation systems need to scale to extremely large graphs of up to billions of edges and target high computational throughput. In contrast, applications such as object detection in point clouds} \cite{shi2020point} or fraud detection \cite{wang2020apan} \hl{rather need to focus on latency and energy efficiency. This highlights the need for acceleration techniques that address not only the challenging GNN computation at relatively small scales and in real time}, but also the storage and multi-GPU coordination issues at larger scales.
\end{enumerate}


\hl{A direct consequence of the aforementioned aspects is that the bottleneck or the critical operation/kernel may vary across GNNs or applications, as shown in} \cite{baruah2021gnnmark, Yan2020a, Zhang2020b}. In light of these challenges, GNNs call for new solutions both in software and hardware. On the software side, several libraries have been proposed to improve the support for GNNs and efficiently compute its multiple variants \hl{both in inference and training. The} extensions of popular libraries such as PyTorch or Tensorflow (TF) \cite{fey2019fast, GNets, grattarola2020graph} \hl{are clear examples of this.} On the hardware side, new accelerator architectures have been surfacing recently \cite{EnGN, AWB, Kiningham2020} that attempt to deal with the flexibility and scalability challenges of GNNs \hl{mostly in inference thus far}. In the next subsections, we provide an exhaustive overview of existing techniques. 



\subsection{Software Frameworks and Accelerators}
The challenges of GNN processing rendered both traditional DNN libraries \hl{and graph processing frameworks} \cite{wang2016gunrock,ham2016graphicionado} inefficient. \hl{The reason is the alternating computing phases of GNNs. DNN libraries would be good at speeding up combination operations within vertices and edges, but perform poorly during aggregation. Graph processing libraries, instead, do a good job at managing irregular memory accesses when traversing the graph. However, these assume trivial operations at the vertices, which is not the case in GNNs}. To bridge this gap, very recent works have started investigating how to adapt the libraries to (i) provide easy to program interfaces to implement multiple GNN variants, (ii) handle the \hl{variety of potentially sparse GNN} operations efficiently in widespread GPU hardware, (iii) scale computations to large-scale graphs and multiple GPUs. 

In the following, we review a comprehensive selection of software frameworks and accelerators, listed in Table \ref{tabsummcomp-SW}. The analysis does not include GunRock \cite{wang2016gunrock} or GE-SpMM \cite{huang2020ge} for different reasons. GunRock, despite implementing GraphSAGE in its latest versions, is a graph processing library \hl{that does not exploit intra-vertex parallelism}. \hl{In fact, two works detailed below} \cite{gnnadvisor,hu2020featgraph} \hl{achieve speedups of} 30$\times$--200$\times$ \hl{with respect to GunRock}. GE-SpMM, although claiming to be tailored to GNNs, is an acceleration method for general-purpose sparse matrix multiplication in GPUs.

A first observation from Table \ref{tabsummcomp-SW} is that software frameworks have been tested for a wide variety of GNN algorithms and relevant datasets. Around 20 different GNN variants have been evaluated, being GCN, GS, and GIN the most common. Even though Amazon, Reddit, Protein, Cora, or CiteSeer datasets are popular in the community, a lack of a widely adopted benchmark suite \cite{hu2020open} makes the datasets to vary widely. It is worth noting, however, that graphs can range from hundreds of edges in chemistry applications to billions of edges in large-scale recommendation systems. \hl{As we see next, performance comparisons are scarce,} but generally take PyG, TF, and DGL as baselines \hl{and often report between one and two orders of magnitude improvement typically in CPU+GPU platforms, with some exceptions on multi-GPU systems} \cite{Jia2019,Ma2019} \hl{or distributed computing clusters with up to 32K cores} \cite{wang2021flexgraph, zhang2020agl}. \hl{Most of the tested frameworks provide optimizations that could work for both acceleration of both training and inference, yet the evaluation is unequal. Training is evaluated in} \cite{fey2019fast,wang2019deep,wang2021flexgraph,zhang2020agl,Jia2019,liu2020g3,hu2020featgraph,Ma2019,Zhu2018ali,Tian2020} \hl{whereas inference time is only measured in} \cite{zhang2020agl,Jia2019,hu2020featgraph,gnnadvisor,Jia2020, greta}.


\begin{table*}
    \centering
    \caption{State of the art in software frameworks and accelerators for GNNs (GS = GraphSAGE)}
    \vspace{-0.4cm}
   \setlength\tabcolsep{3pt}
    \small
    \begin{tabular}{|>{\centering\arraybackslash}m{1.2cm}|>{\centering\arraybackslash}m{7cm}|>{\centering\arraybackslash}m{1.6cm}|>{\centering\arraybackslash}m{2.8cm}|>{\centering\arraybackslash}m{1.2cm}|}  \hline
         
         \multirow{2}{*}{\textbf{Name}} & \multirow{2}{*}{\textbf{Main Features}} & \multicolumn{3}{c|}{\textbf{Evaluation}} \\ \cline{3-5}
           &  &  \textbf{Algorithms} & \textbf{Datasets} & \textbf{Baselines} \\ \hline
         
         PyG \cite{fey2019fast}
            & \begin{itemize}[leftmargin=5pt]
                \item Leverages widespread adoption of PyTorch.
                \item Wide variety of example codes available.
                \item Use of scatter-gather kernels + node/edge parallelism. \item Evaluated in GPU. \hl{Compatible with BigGraph} \cite{Lerer2019} \hl{to scale.} \vspace{-0.35cm}
            \end{itemize}%
            & GCN, GAT, SGC, GS, GIN, etc...%
            & Cora, CiteSeer, PubMed, MUTAG, Proteins, Collab, IMDB, Reddit%
            & DGL \\ \hline%
        
        
               DGL \cite{wang2019deep}
            & \begin{itemize}[nosep,after=\strut,leftmargin=5pt]
                \item Library compatible with TF, PyTorch and MXNet.
                \item Deep documentation and support, tutorials.
                \item Based on matrix-mul kernels. Evaluation in CPU and GPU.
                \item \hl{Augmented with DistDGL} \cite{zheng2020distdgl} \hl{for distributed computing.}\vspace{-0.35cm}
            \end{itemize}  
            & GCN, GAT, SGC, GS, GIN, R-GCN, GCMC
            & Reddit, OGB (Arxiv, Protein, Product, Citation, PPA), Movielens
            & PyG \\ \hline
         
        NeuGraph \cite{Ma2019} 
            & \begin{itemize}[nosep,after=\strut,leftmargin=5pt]
                \item \vspace{0.1cm}Implementation and evaluation for scaling to multiple GPUs.
                \item Four-function model allowing for updates at edges and nodes.
                \item Optimized partitioning, scheduling, pipelining, transfers.
                \item Built on TF, not open sourced. \vspace{-0.3cm}
            \end{itemize}  
            & GCN, CommNet, GG-NN
            & Pubmed, Blog, Reddit, Enwiki, Amazon  
            & DGL, TF \\ \hline

         AliGraph \cite{Zhu2018ali} &  
         \begin{itemize}[nosep,after=\strut,leftmargin=5pt]
                \item \vspace{0.1cm}Targeting large-scale graphs and distributed systems. 
                \item Emphasis on distributed storage and partitioning
                \item Only work with heterogeneous and dynamic GNNs, and huge datasets (up to 483M edges, 6.5B edges). Built on top of TF.\vspace{-0.3cm}
            \end{itemize}
            & GS, six in-house algorithms   
            & Amazon, Taobao  
            & N/A \\ \hline 
         
           \hl{FlexGraph} \cite{wang2021flexgraph} &  \begin{itemize}[nosep,after=\strut,leftmargin=5pt]
                \item \vspace{0.1cm}\hl{Uses NAU programming model for flexible aggregation.}
                \item \hl{Hierarchical aggregation with dynamic sparse-dense logic. }
                \item \hl{Supports distributed computing, tested in 1500-core system.}  \vspace{-0.3cm}
            \end{itemize}-
            & \hl{GCN, PinSage, MAGNN}
            & \hl{Reddit, FB91, Twitter, IMDB}
            & \hl{PyG, DGL, DistDGL, Euler} \\ \hline

         AGL \cite{zhang2020agl} &  
         \begin{itemize}[nosep,after=\strut,leftmargin=5pt]
                \item \vspace{0.1cm}Aiming for scalability, fault tolerance, and integrality.
                \item \hl{Uses MapReduce to scale, tested in 32000-core system.}\vspace{-0.3cm}
            \end{itemize}
            & GCN, GS, GAT
            & Cora, PPI, UUG,
            & PyG, DGL \\ \hline 
                   
                ROC \cite{Jia2019} &  
         \begin{itemize}[nosep,after=\strut,leftmargin=5pt]
                \item \vspace{0.1cm}Implemented on top of FlexFlow \cite{jia2019beyond}.
                \item Optimizations: dynamic partitioning, memory management.
                \item Evaluated with single and multiple GPUs via NVLink.\vspace{-0.3cm}%
            \end{itemize}%
            &  GCN, GS, CommNet, GIN, FastGCN
            & Pubmed, PPI, Reddit, Amazon  
            & TF, DGL, PyG, NeuGraph \\ \hline 
            
            
            GNN Advisor \cite{gnnadvisor} &  \begin{itemize}[nosep,after=\strut,leftmargin=5pt]
                \item \vspace{0.1cm}Unique runtime profiling of graph information (degree, feature size, communities) to guide GPU processing
                \item Extensive comparison with similar frameworks in single GPU\vspace{-0.3cm}%
            \end{itemize}%
            & GCN, GIN%
            & CiteSeer, Cora, Pubmed, PPI, Prot, Yeast, DD,twit, SW620H, amazon, artist
            & DGL, PyG, GunRock, NeuGraph \\ \hline%
            
                    
            

         
         PCGCN \cite{Tian2020}  & 
         \begin{itemize}[nosep,after=\strut,leftmargin=5pt]
                \item \vspace{0.1cm}Motivated by power-law distribution of node degrees.
                \item Optimized partitioning to generate dense matrices.
                \item Dual execution mode depending on sparsity of each partition.
                \item Built on top of TF, evaluated in single GPU. \vspace{-0.3cm}
            \end{itemize}
            & GCN 
            & Pubmed, Blog, Youtube, C1000-9, MANN-a81, Reddit, synthetic (RMAT) 
            & TF, DGL, PyG \\ \hline
         
         
         HAG \cite{Jia2020} &  \begin{itemize}[nosep,after=\strut,leftmargin=5pt]
                \item \vspace{0.1cm}Removes redundant sums in aggregation by \emph{fusing} nodes.
                \item Runtime algorithm to \emph{fuse} nodes only if predicted beneficial.
                \item The impact on operation reduction is independent of hardware, but the impact on execution speed is not.\vspace{-0.3cm}
            \end{itemize}
            & GCN, GIN, SGC   
            & BZR, PPI, Reddit, IMDB, COLLAB
            & N/A \\ \hline
         
         
         FeatGraph \cite{hu2020featgraph} &  \begin{itemize}[nosep,after=\strut,leftmargin=5pt]
                \item \vspace{0.1cm}Optimized matmul kernels for aggregation and combination.
                \item User-defined combination functions and optimizations.\vspace{-0.3cm}
            \end{itemize}  
            & GCN, GS, GAT
            & OGB (Proteins), Reedit, sythetic graphs
            & GunRock \\ \hline
            
        \hl{G}\textsuperscript{3} \cite{liu2020g3} &  \begin{itemize}[nosep,after=\strut,leftmargin=5pt]
                \item \vspace{0.1cm}\hl{Brings together graph processing frameworks and GNNs.}
                \item \hl{Offers APIs over C/C++ for ease of programming.}
                \item \hl{Uses GunRock} \cite{wang2016gunrock} \hl{to provide GPU runtime optimizations.}  \vspace{-0.3cm}
            \end{itemize} 
            & \hl{GCN, SGC}
            & \hl{PubMed, Reddit}
            & \hl{PyG, TF} \\ \hline
         
         GReTA \cite{greta} & \begin{itemize}[nosep,after=\strut,leftmargin=5pt]
                \item \vspace{0.1cm}Programming abstraction with user-defined functions, similar to SAGA, targeting accelerators and any GNN variant.
                \item Evaluation based on GRIP (see Table \ref{tabsummcomp-HW}) \hl{in ASIC.}\vspace{-0.3cm}
            \end{itemize}  & GCN, GS, G-GCN, GIN
            & Youtube, Livejournal, Pokec, Reddit 
            & N/A \\ \hline
    \end{tabular}
    \label{tabsummcomp-SW}
    \vspace{-0.2cm}
\end{table*}

\vspace{0.06cm} \noindent 
\textbf{\underline{PyTorch Geometric (PyG).}} PyG \cite{fey2019fast} is a widespread library that is built upon PyTorch and that provides support for relational learning, illustrated in a myriad of algorithms. The key aspect is the definition of a message passing interface with definition of \texttt{message} and \texttt{update} functions for neighbourhood aggregation and combination, respectively, and multiple pooling operations. To accelerate GNN processing, PyG handles sparsity via dedicated GPU scatter and gather kernels that operate in all edges and nodes in parallel, instead of using sparse matrix multiplication kernels. \hl{Relevantly, Facebook released Pytorch-BigGraph} \cite{Lerer2019}, a library that allows to process arbitrarily large graphs by introducing partitioning and distributed processing and that could complement PyG.

\vspace{0.06cm} \noindent 
\textbf{\underline{Deep Graph Library (DGL).}} DGL \cite{wang2019deep} is a recent library that works on top of TF, PyTorch, or MXNet, and provides plenty of examples and code for multiple GNNs. The library defines three functions: \texttt{message} for edge aggregation and update and \texttt{reduce} and \texttt{update} for aggregation and combination at the nodes. To boost performance, DGL takes a matrix multiplication approach and leverages specialized kernels for GPUs or TPUs. In particular, both sampled dense-dense and sparse matrix multiplications are considered together with node, edge or feature parallelization. As discussed in their work \cite{wang2019deep}, DGL uses heuristics to choose among the different options as the optimal parallelization scheme depends on multiple factors including the input graph. Thanks to this approach, \hl{DGL claims to achieve an order of magnitude faster training than PyG. Recently, researchers at Amazon have released a DistDGL, a system based on DGL for distributed mini-batch training scalable to billion-edge graphs} \cite{zheng2020distdgl}. \hl{To achieve it, DistDGL uses min-cut graph partitioning via a lightweight algorithm.}

\vspace{0.06cm} \noindent 
\textbf{\underline{NeuGraph.}} Microsoft Research led one of the first specialized frameworks for parallel processing of \acrshort{gnnpl} in \acrshort{gpu}s, NeuGraph \cite{Ma2019}. Although it is built on top of TF, NeuGraph is not open source at the time of this writing. The framework implements a programming model, SAGA-NN, based on the functions \texttt{Scatter} for edge aggregation, \texttt{ApplyEdge} for edge combination, \texttt{Gather} for node aggregation, and \texttt{ApplyVertex} for node combination. 
Scatter-gather kernels are used in the functions of the same name, whereas matrix multiplication primitives are used in the combination functions. NeuGraph also features a number of optimizations to accelerate GNN computing. First, the partitioning of large graphs performed via the Kernighan-Lin algorithm to make partitions denser and minimize the transfers between partitions, which harm performance. Second, scheduling of partitions to the GPU is optimized by batching together small sparse partitions that can be computed together \cite{Nagasaka2019}, and also profiling transfer and computation times in first GNN layer to later pipeline different chunks perfectly. Third, NeuGraph also eliminates redundant computation by fusing multiple edges together. Finally, it allow to scale GNN to multiple GPUs by distributing the computation, and optimizes the transfer of information by using a ring-based dataflow that minimizes contention at the interconnect.

\vspace{0.06cm} \noindent 
\textbf{\underline{AliGraph.}} Developed by the AliBaba group and open-sourced with the name of \texttt{graph-learn}, 
AliGraph is a GNN framework built on top of TF \cite{Zhu2018ali}. The framework is thought for the processing of very large and dynamic graphs in large-scale computing systems, and is currently used in recommendation services at AliBaba. It implements three layers, namely: \textit{storage}, that implements partitioning with four different algorithms, but in this case to store the graph in a distributed way; \textit{sampling}, which unlike other frameworks, allows to define custom sampling of a nodes' neighbourhood relevant to algorithms such as GraphSAGE; and \textit{operator}, which implements the aggregation and combination functions. In overall, the AliGraph is unique due to its distributed approach and the many optimizations made at the storage layer to minimize data movement, such as the use of four different partitioning algorithms depending on the characteristics of the graph, or caching important vertices in multiple machines to reduce long misses.

\vspace{0.06cm} \noindent 
\textbf{\underline{FlexGraph.}} The AliBaba group also leads the development of FlexGraph \cite{wang2021flexgraph}, \hl{a distributed framework for GNN training whose distinct features are their flexible definitions of neighbourhood and the hierarchical aggregation schemes. To this end, FlexGraph uses the NAU programming model described in Section} \ref{sec:algorithms}. \hl{To speedup training, FlexGraph combines hierarchical aggregation with a hybrid execution strategy combining sparse and dense logic. It also accelerates distributed execution through an application-driven workload balancing strategy and a pipeline processing strategy to overlap computations and communications.}


\vspace{0.06cm} \noindent 
\textbf{\underline{AGL.}} AGL \cite{zhang2020agl} is a framework created specifically for industral deployments of massive GNNs. To that end, the authors emphasize their scalability, fault tolerance, and use of existing widespread methods for distributing the computation. In particular, AGL uses MapReduce \cite{dean2008mapreduce} to that end and tests the proposed system in CPU clusters. The framework has three modules: one for creating independent neighbourhoods that can be processed in parallel, one for optimizing training, and one for the slicing of the graph and calculation of inference. Numerous optimizations are proposed in the sampling and indexing of the graph, partitioning and pruning, and pipelining of computation during training.  

\vspace{0.06cm} \noindent 
\textbf{\underline{ROC.}} ROC \cite{Jia2019} is another GNN framework targeting multi-GPU systems, in this case built on top of FlexFlow \cite{jia2019beyond}. Similarly to AliGraph or AGL, ROC is able to distribute large graphs to multiple machines. However, this framework differs from others in that the partitioning method and memory management is performed with dynamic methods providing extra acceleration. First, ROC uses an online linear regression model to approach partitioning optimally. This model uses the training iterations to learn the best strategy of a specific graph, outperforming static methods significantly. Second, memory management is treated as a cost minimization problem and solved via an online algorithm that finds where to best store each partition. The authors demonstrate that such acceleration methods provide better scalability than DGL and PyG in single GPUs, and better scaling to multiple GPUs than NeuGraph.

\vspace{0.06cm} \noindent 
\textbf{\underline{GNNAdvisor.}} The work by Wang \emph{et al.} \cite{gnnadvisor} presents a runtime system that aims to systematically accelerate GNNs on GPUs. Instead of treating this problem via abstract models as done in ROC, GNNAdvisor does an online profiling of the input graph and GNN operations to guide the memory and workload management agents at the GPU. In particular, it leverages (i) the node degree to fine-tune the group-based workload management of the GPU, (ii) the size of the node embedding to optimize workload sharing, and (iii) the existing of communities within the graph to guide partitioning and scheduling. While the two first features are trivial to obtain, community detection is generally harder. In this case, the authors use a combination of node renumbering and Reverse Cuthill–McKee algorithm to reorder the adjacency matrix in a way that dense partitions are available. \hl{Thanks to all these techniques, the authors claim} 3$\times$-4$\times$ speedup over DGL, PyG, and NeuGraph in a high-end GPU. 

               
\vspace{0.06cm} \noindent 
\textbf{\underline{PCGCN.}} The paper by Tian and co-authors \cite{Tian2020} present a partition-centric approach to acceleration of GNNs in GPUs, which they implement on top of TF. The contribution is motivated by the power-law distribution of the node degrees in a graph, which largely affects partitioning. PCGCN applies a locality-aware partitioning, METIS \cite{METIS}, that helps obtaining dense sub-matrices. That, however, does not prevent sparse partitions to appear. To combat this, PCGCN profiles the partitions at runtime and applies a dual-mode of operation: dense matrix representation and multiplication kernels when dense, and column-sparse representation and sparse kernels otherwise. In the paper, the authors compare their implementation with vanilla TF, and also DGL and PyG, and report the lowest speedup across libraries. \hl{Even in this case, PCGCN always speeds up execution and achieves upto 8.8}$\times$ in highly clustered graphs.

\vspace{0.06cm} \noindent 
\textbf{\underline{HAG.}} This work presents the concept of \emph{Hierarchically Aggregated computation Graph} (HAG) \cite{Jia2020}. The authors make the observation that many of the operations made during the aggregation stage are repeated multiple times when nodes share similar neighbourhoods. In response to this, HAGs are presented as an alternative representation that proactively ``fuses" nodes with common neighbourhoods, removing redundant aggregations during the execution of any GNN. Since the search of similarly-connected nodes can be expensive, HAG employs a cost function to estimate the cost of certain node fusions, to then adopt a search algortihm affordable for runtime. With only 0.1\% of \hl{memory overhead, HAG reduces the amount of aggregations by 6.3}$\times$.

\vspace{0.06cm} \noindent 
\textbf{\underline{FeatGraph.}} Developed in collaboration with Amazon, FeatGraph \cite{hu2020featgraph} proposes to optimize kernels of aggregation and combination separately. Different from other frameworks, here the user can define the combination function and ways to parallelize it, so that the scheduler can take it into account. As optimizations, FeatGraph also proposes to combine graph partitioning with feature dimension tiling and to adopt a hybrid partitioning scheme for GPUs.

\vspace{0.06cm} \noindent 
\textbf{\underline{G\textsuperscript{3}.}} Liu \emph{et al.} \cite{liu2020g3} \hl{propose a framework for the training of GNNs in GPU systems. G\textsuperscript{3} facilitates the task of GNN creation by providing a set of flexible APIs over C/C++ code that implement widespread layers and models. G\textsuperscript{3} also incorporates a set of graph-centric optimizations based on GunRock for aggregation} \cite{wang2016gunrock} \hl{dealing with memory management, workload mapping, and load balancing. In training, G\textsuperscript{3} shows up to 100X speedup over PyG and TF in a high-end GPU}.

\vspace{0.06cm} \noindent 
\textbf{\underline{GReTA}} GReTA \cite{greta} is a processing abstraction for \acrshort{gnn}s aiming at simplifying their representation for hardware implementations. 
To this end, GReTA consists of four user-defined functions: \texttt{Gather} and \texttt{Reduce} to describe the aggregation, and \texttt{Transform} and \texttt{Activate} to describe the combination. These functions enable certain flexibility to accommodate different \acrshort{gnn} types. GReTA also discusses partitioning briefly and exemplifies it in a hardware accelerator called GRIP \cite{Kiningham2020}, which is described in the next section.

\vspace{0.06cm} \noindent 
\textbf{\underline{Paddle Graph Learning (PGL).}} Developed by Baidu Research, PGL \cite{PGL} \hl{is a graph learning framework based on PaddlePaddle} \cite{ma2019paddlepaddle} \hl{that supports both walk-based and message passing models in heterogeneous graphs. Moreover, it integrates a Model Zoo supporting many GNN models to foster adoption, as well as support for distributed computing.} 

\vspace{0.06cm} \noindent 
\textbf{\underline{Tripathy \emph{et al.}}} 
In this work, the authors compare multiple parallelization algorithms that partition and distribute the \acrshort{gnn} in multiple GPU clusters, i.e., 1D, 1.5D, 2D and 3D algorithms, and model the tradeoff between inter-GPU communication and memory requirements of these setups analytically and for training. The model takes a large adjacency matrix and breaks it down to a fixed amount of processes depending on the algorithm. Then, an analysis is made on the amount of effectual operations and results to be communicated across the GPUs. Their implementation over PyG shows promising scalability and nominates the 1.5-D algorithm as a promising and balanced alternative, although the best algorithm depends on the characteristics of the input graph.

\subsection{Hardware Accelerators}
\label{sec:HW}
We have seen above that software accelerators streamline the execution of GNNs in CPU-GPU platforms present in most computing systems, achieving significant speedups \hl{both in inference and training}. Fewer works \cite{Balog2019, Zhang2020b} have tested GNN training in the TPUs typically used in \emph{dense} DNNs, showing similar performance than in GPUs. 

\begin{figure*}
    \centering
    \includegraphics[width=0.95\textwidth]{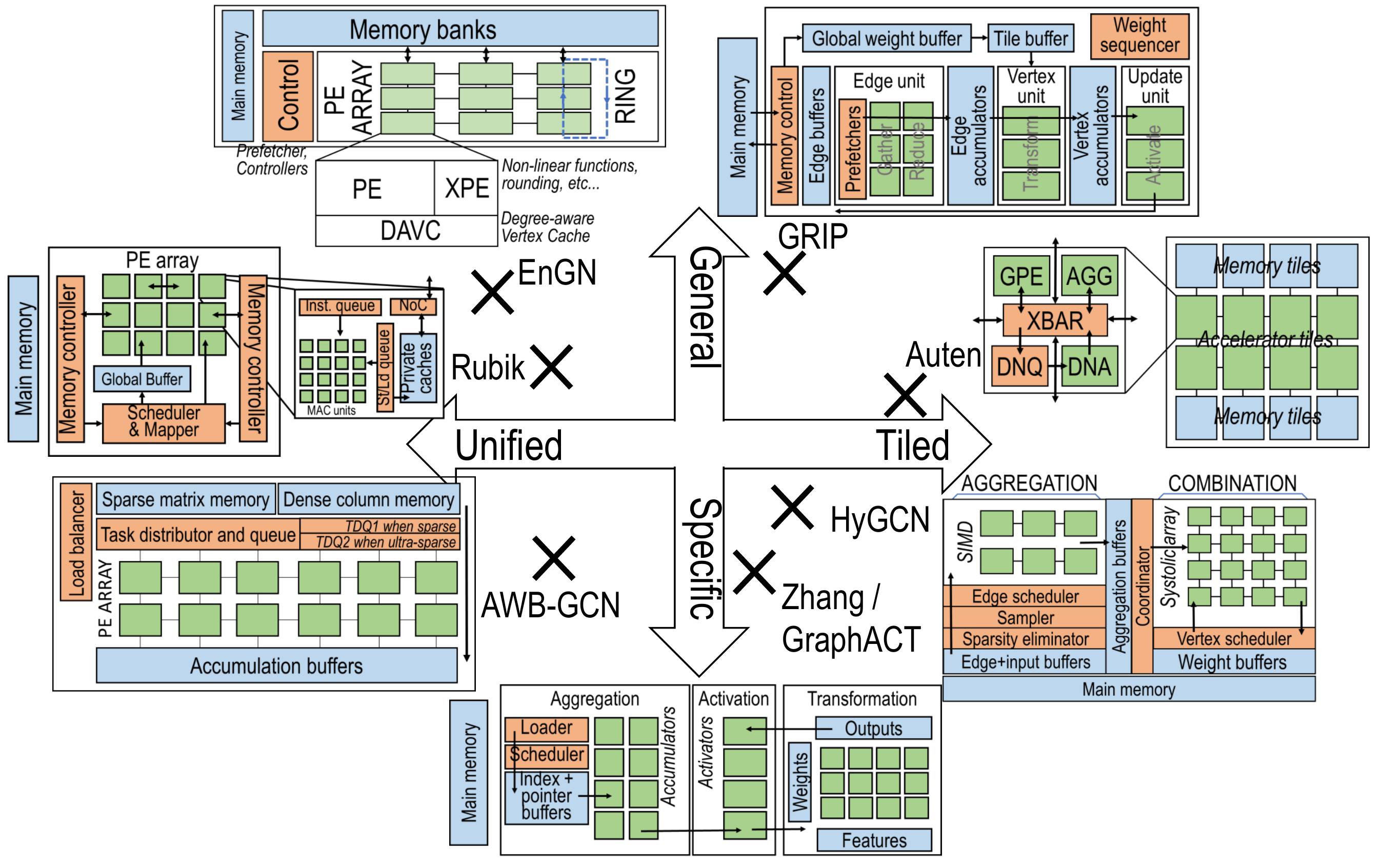}
    \vspace{-0.5cm}
    \caption{\hl{Qualitative classification} and schematic representation of hardware accelerators for GNN inference. Green, blue, and red squares represent processors, memory, and control units, respectively.}
    \vspace{-0.4cm}
    \label{accel}
\end{figure*}

In this context, a pertinent question is whether custom hardware accelerators can tackle the unique challenges of GNN computing and live up to the promise of order-of-magnitude improvements that, to cite an example, have been already achieved in CNNs \cite{eyeriss2}. Pursuing this goal, several hardware accelerators have emerged which attempt to handle the extreme density and alternating computing requirements of GNNs. We next discuss all the designs published to date, using as reference the schematic diagrams of their architecture shown in Fig. \ref{accel}. The figure also tries to classify the architectures in two axes: \hl{unified versus tiled to assess whether the computing phases are physically separated and how tightly coupled they are; and general to specific to assess} how easy is to adapt the accelerator to multiple GNN variants.

A summary of the main features of the accelerators and evaluated algorithms and datasets is given in Table \ref{tabsummcomp-HW}. We observe that most works revolve around the GCN algorithm, which is popular and easy to illustrate. Datasets are generally smaller than in software acceleration works, mainly because of the memory limitations of hardware accelerators in inference and the cost of simulating hardware architectures. Cora, CiteSeer, and Reddit are the most common ones. While performance comparisons are difficult due to the many variables involved, most works use CPUs and GPUs as baselines and, in some cases, \hl{even HyGCN} \cite{Yan2020} \hl{and AWB-GCN} \cite{AWB} \hl{as early works on hardware acceleration. In general, the proposed accelerators are around two and three orders of magnitude faster and more energy efficient than GPUs and CPU platforms, respectively, often occupying less than 10 mm}\textsuperscript{2}. There is no consensus on which software framework shall be used in the baselines. \hl{Finally, all accelerator proposals except GraphACT are designed and evaluated for inference.}

\vspace{0.06cm} \noindent 
\textbf{\underline{EnGN.}} Among the first accelerators to appear, EnGN \cite{EnGN} presents a unified architecture heavily inspired by CNN accelerators.
The GNN is fundamentally treated as concatenated matrix multiplication of feature vectors, adjacency matrices, and weights --all scheduled in a single dataflow. An array of clustered Processing Elements (PEs) is fed by independent banks for the features, edges, and weights to compute the combination function. To perform the aggregation, each column of PEs is interconnected through a ring and results are passed along and added according to the adjacency matrix in a process the authors call Ring-Edge Reduce (RER). Within this architecture, sparsity is handled with several optimizations. First, the RER aggregation may lead to multiple ineffectual computations for sparsely connected nodes. To avoid this, EnGN reorders edges on the fly in each step of the RER. Second, PE clusters are attached to a degree-aware vertex cache that holds data regarding high-degree vertices. The reasoning is that well-connected vertices will appear multiple times during the computation and caching them will provide high benefit at modest cost. Other optimized design decisions relate to the order of the matrix multiplications when the aggregation function is sum, which affects the total number of operations, or the tiling strategy, which affects data reuse and I/O cost.


\begin{table*}
    \centering
    \caption{State of the art in hardware accelerators for GNNs.}
    \vspace{-0.4cm}
    \setlength\tabcolsep{3pt}
    \small
    \begin{tabular}{|>{\centering\arraybackslash}m{1.2cm}|>{\centering\arraybackslash}m{7.2cm}|>{\centering\arraybackslash}m{1.5cm}|>{\centering\arraybackslash}m{2.4cm}|>{\centering\arraybackslash}m{1.6cm}|}  \hline
        
   \multirow{2}{*}{\textbf{Name}} & \multirow{2}{*}{\textbf{Main Features}} & \multicolumn{3}{c|}{\textbf{Evaluation}} \\ \cline{3-5}
           &  &  \textbf{Algorithms} & \textbf{Datasets} & \textbf{Baselines} \\ \hline

        EnGN \cite{EnGN} 
        &  \begin{itemize}[nosep,after=\strut,leftmargin=5pt]
            \item \vspace{0.1cm}Unified architecture with dense hardware, single dataflow, generalizable to many GNN variants.
            \item Aggregation via Ring-Edge Reduction (RER).
            \item Optimizations: edge reordering, degree-aware vertex cache, scheduling.\vspace{-0.3cm}
        \end{itemize} 
        & GCN, GS, GG-NN, GRN, R-GCN 
        & Cora, PubMed, Nell, Reddit, Enwiki, Amazon, synthetic (RMAT), AIFB, MUTAG, BGS, AM   
        & CPU-DGL, GPU-DGL, CPU-PyG, GPU-PyG, HyGCN  \\ \hline

         HyGCN \cite{Yan2020}%
         &  \begin{itemize}[nosep,after=\strut,leftmargin=5pt]
             \item \vspace{0.1cm}Hybrid architecture with separate aggregate/combine phases.
             \item Fine-grained pipelining via inter-phase coordinator.
             \item Eliminates sparsity with window sliding/shrinking approach.
             \item Focused on GCNs, unclear how to generalize (no edge updates).\vspace{-0.3cm}%
         \end{itemize} 
        & GCN, GSC, GIN, DiffPool%
        & IMDB, Cora, CiteSeer, COLLAB, PubMed, Reddit%
        & CPU-PyG, GPU-PyG  \\ \hline
         
          AWB-GCN \cite{AWB} 
         &  \begin{itemize}[nosep,after=\strut,leftmargin=5pt]
             \item \vspace{0.1cm}Adapts to varying GNN workloads via three load balancing techniques, \hl{chosen based on the sparsity of each partition.}
             \item \hl{Processes combination first to reduce the number of operations.}
             \item \hl{Fine-grained pipelining of aggregation and combination.}
             \item Focused on GCNs, unclear how to generalize.\vspace{-0.3cm}
         \end{itemize} 
         & GCN  
         & Cora, CiteSeer, PubMed, Reddit, Nell
         & CPU-PyG, GPU-PyG, FPGA, HyGCN  \\ \hline

        GRIP \cite{Kiningham2020} 
         & \begin{itemize}[nosep,after=\strut,leftmargin=5pt]
                \item \vspace{0.1cm}Uses the GReTA abstraction \cite{greta}, generalizable to any GNN.
                \item Actual implementation with techniques similar to HyGCN.\vspace{-0.3cm}
            \end{itemize} 
            & GCN, GIN, G-GCN, GS%
            & Youtube, Livejournal, Pokec, Reddit%
            & CPU-TF, GPU-TF, \hl{TPU, HyGCN}  \\ \hline

         Auten \emph{et al.} \cite{Auten2020} 
         & \begin{itemize}[nosep,after=\strut,leftmargin=5pt]
                 \item \vspace{0.1cm}Tiled architecture, ready for scale-out via Network-on-Chip.
                 \item Similar to HyGCN, less specialized but easier to generalize.\vspace{-0.3cm}
                 \end{itemize} 
                 & GCN, MPNN, GAT, PGNN%
                 & Cora, CiteSeer, DBLP, PubMed, QM9\_1000%
                 & CPU, GPU   \\  \hline
         
         Zhang \emph{et al.} \cite{Zhang2020acc} & 
         \begin{itemize}[nosep,after=\strut,leftmargin=5pt]
                 \item \vspace{0.1cm}Combination of offline software acceleration (redundancy elimination + node reordering) and hardware acceleration in FPGA.
                 \item Optimizations: double buffering, node+feature parallelism, dual pipelining mode depending of order of matrix multiplications.\vspace{-0.3cm}
                 \end{itemize} 
                 & GCN
                 & Flickr, Reddit, Yelp  
                 & CPU-TF, GPU-TF, CPU-C++, GPU-C++ \\ \hline 
       
       \hl{Rubik} \cite{chen2020rubik}  & 
       \begin{itemize}[nosep,after=\strut,leftmargin=5pt]
                 \item \hl{Hierarchical and unified PE array design}
                 \item \hl{Includes small caches to eliminate redundant aggregations}
                 \item \hl{Adds graph reordering in software to improve cache utilization}\vspace{-0.3cm}%
                 \end{itemize} 
                 & \hl{GIN, GS}%
                 & \hl{Collab, BZR, IMDB, DD, CiteSeer, Reddit}%
                 & \hl{Eyeriss-like, GPU-PyG}\\ \hline
        
        \hl{GCNAX} \cite{li2021gcnax}  & 
       \begin{itemize}[nosep,after=\strut,leftmargin=5pt]
                 \item \hl{Architecture with reconfigurable loop ordering and fusion.}
                 \item \hl{Choice is made after an offline design space exploration.}
                 \item \hl{Uses outer product to mitigate unbalanced presence of zeros.}\vspace{-0.3cm}%
                 \end{itemize} 
                 & \hl{GCN}%
                 & \hl{Cora, CiteSeer, Pubmed, Nell, Reddit}%
                 & \hl{HyGCN, AWB-GCN, SpArch}\\ \hline

       GraphACT \cite{Zeng2020}  & 
       \begin{itemize}[nosep,after=\strut,leftmargin=5pt]
                 \item \vspace{0.1cm}Only accelerator evaluating training and memory footprint.
                 \item CPU+FPGA. Optimizations rely on load balancing, scheduling, batching, removal of redundant aggregation operations.\vspace{-0.3cm}
                 \end{itemize} 
                 & GCN%
                 & PPI, Reddit, Yelp%
                 & CPU, GPU \\ \hline

    \end{tabular}
    \vspace{-0.3cm}
    \label{tabsummcomp-HW}
\end{table*}

\vspace{0.06cm} \noindent 
\textbf{\underline{HyGCN.}} The authors HyGCN \cite{Yan2020} build upon the observation that GNNs present two main alternating phases of opposed computation needs, to introduce a hybrid architecture for \acrshort{gcn}s. HyGCN is composed of separate dedicated engines for the aggregation and the combination stages, plus a control mechanism that coordinates the pipelined execution of both functions. Being dense, the combination stage is computed via a conventional systolic array approach. The aggregation stage has a more elaborated architecture featuring a sampler, an edge scheduler, and a sparsity eliminator that feeds a set of SIMD cores. Within this architecture, sparsity is handled at the aggregation engine thanks to efficient scheduling and the sparsity eliminator. The latter takes a window-based sliding and shrinking approach to dynamically adapt to varying degrees of sparse multiplications. To further adapt to the workloads, HyGCN allows to group the SIMD cores in aggregation and the PEs in combination in different ways depending on the size of feature vectors. Finally, special attention is placed to the design of the inter-engine coordinator to optimize memory accesses and allow fine-grained pipelining of the execution towards maximizing parallelism dynamically. 

\vspace{0.06cm} \noindent 
\textbf{\underline{AWB-GCN.}} The \emph{Autotuning-Workload-Balancing} GCN accelerator \cite{AWB} advocates for an aggressive adaptation to the structural sparsity of the \acrshort{gnn}. The authors motivate their design by analyzing the power-law distribution of most graphs, arguing that some parts of the computation will be dense and others extraordinarily sparse, creating unbalances. To address the imbalance, the architecture develops a custom matrix multiplication engine with efficient support of skipping zeros. To that end, data from memory is fed via a task distributor and queue (TDQ) to a set of PEs and accumulators. The TDQ takes two designs adapted to when sparsity is moderate or high. \hl{Since AWB-GCN focuses on GCNs which have linear aggregation functions, the authors propose to process combination first as this generally reduces the amount of features and, thus, the amount of operations performed in aggregation. Furthermore, AWB-GCN provides a fine-grained pipelining mechanism to overlap the execution of combination and aggregation even within the same layer. However,} the key of AWB-GCN are its three workload balancing functions. The first one is local and tries to balance the load among neighboring PEs. The second one is remote and attempts to pour overflowing computation from a busy PE to a single remote underutilized PE. The third one takes the load of extremely busy PEs processing very dense node clusters and divides across multiple idle PEs. To support that, AWB-GCN provisions hardware at the TDQ and the connections to the PEs to allow the remapping of nodes to remote PEs and to take them back for coherent aggregation. Moreover, all decisions are taken based on information extracted from simple counting at the queues.


\vspace{0.06cm} \noindent 
\textbf{\underline{GRIP.}} A key aspect of most existing accelerators is that they focus on GCNs as a relevant GNN algorithm. In contrast, the GRIP accelerator \cite{Kiningham2020} leverages the abstraction of GReTA \cite{greta} to develop a general accelerator for any GNN variant, allowing to perform edge and node updates with user-defined functions. The GRIP architecture reflects this by having separated and custom units and accumulators for both edges (gather, reduce) and vertices (transform, activate). A control unit orchestrates data movement between the different units and respective buffers. In the sample implementation, GRIP divides the edge update unit into lanes to execute vertices simultaneously and takes an input-stationary dataflow for the vertex update unit. Among the optimizations made, we found pipelining and tiling adapted to the particularities of the implemented dataflows, similar to that of other accelerators.

\vspace{0.06cm} \noindent 
\textbf{\underline{Auten \emph{et al.}}} Unlike most other accelerators, this work \cite{Auten2020} proposes a modular architecture for convolutional GNNs. The basic unit of the accelerator is a tile composed by an aggregator module (AGG), a \acrshort{dnn} accelerator module (DNA), a \acrshort{dnn} queue (DNQ) and a graph \acrshort{pe} (GPE), all of them connected to an on-chip router. Thus, the architecture can be scaled out by interconnecting multiple tiles among them and with memory. Within each tile, the architecture has a similar structure than HyGCN, with the DNA being an array for dense multiplication, the AGG an edge-controlled adder, the DNQ taking the role of inter-engine buffer, and the GPE controlling execution. In this case, however, the GPE is a lightweight CPU managing multiple threads rather than an optimized controller.

\vspace{0.06cm} \noindent 
\textbf{\underline{Zhang \emph{et al.}}} The work by Zhang and co-authors \cite{Zhang2020acc} presents a combination of software and hardware acceleration for GCNs. On the one hand, the graph is pre-processed via a redundancy elimination mechanism similar to that of \cite{Jia2020} and a node reordering similar to that of \cite{gnnadvisor}. Pre-processing is done offline and is justified for the repeated benefits that it can provide to multiple inferences to static graphs. The processed graph is then fed to a hardware accelerator implemented in a FPGA consisting of differentiated pipelined modules for aggregation (sparse array) and combination (dense systolic array and non-linear activation module). As differentiating elements with respect to other designs, we find that the aggregator module uses a double-buffering technique to hide latency of additions, and exploits both node-level and feature-level parallelism. We also observe that the accelerator implements two modes of operation depending on the order of the matrix multiplications, which leads to different strategies for pipelining. To accommodate them, the modules are interconnected both from the aggregate module to the combination modules, and \emph{vice versa}.

\vspace{0.06cm} \noindent 
\textbf{\underline{Rubik.}} \hl{Similar to the case above, Rubik} \cite{chen2020rubik} \hl{proposes a hardware accelerator assisted by some pre-processing in software. On the hardware side, Rubik presents a hierarchical PE array design, wherein each PE contains a number of MAC units plus instruction and data queues to feed them. The design is unified because aggregations and combinations are scheduled across all PEs. Moreover, each PE includes two small private caches that store recently accessed vertices and partial aggregations. Each PE is connected to the rest of PEs and two memory controllers placed on the side via a meshed NoC. On the software side, Rubik proposes lightweight graph reordering (once per graph) to put together nodes that are connected with each other, similarly to} \cite{gnnadvisor}\hl{, but here to improve the performance of the private PE caches.}

\vspace{0.06cm} \noindent 
\textbf{\underline{\hl{GCNAX.}}} \hl{The work in} \cite{li2021gcnax} \hl{points out the load imbalance, execution order, and loop optimization inefficiencies from other accelerators, whose impact varies across workloads. To address them, the authors propose GCNAX as a flexible accelerator whose dataflow is reconfigurable in terms of loop order and loop fusion strategy. To find the most effective dataflows for each particular dataset, the authors perform a design space exploration of dataflow design decisions. Therefore, in inference, GCNAX is reconfigured based on the characteristics of the problem at hand. Finally, GCNAX uses the outer product to mitigate the effect of unbalanced presence of zeros, unlike other accelerators. Thanks to these techniques, GCNAX is around 10}$\times$ and 2$\times$ faster and more efficient than HyGCN and AWB-GCN, respectively.

\vspace{0.06cm} \noindent 
\textbf{\underline{GraphACT.}} While all other accelerators focused on inference, GraphACT \cite{Zeng2020} explores \hl{how to efficiently perform GNN training in an heterogeneous CPU+FPGA platform.} The main design decision relates to determining which parts are computed where and which data to store in memory. To address these questions, the authors argue that CPU performs graph sampling and the calculation of the loss gradients, while and the FPGA does forward and backward propagation passes. The FPGA, thus implements aggregation and combination. The authors present optimizations based on the scheduling of the different operations taking into consideration that backpropagation can be performed after batching of multiple layers or batching different parts of the graph. Moreover, similarly to in \cite{Zhang2020acc}, redundant operations at aggregation are eliminated via searching of edges common to multiple vertices.



\subsection{Discussion}
\hl{The analysis of the state of the art performed in previous sections leads to several conclusions. First, we observe that a quantitative comparison among systems is very difficult due to the lack of a common baseline system and a GNN benchmark suite with a representative set of algorithms, datasets, and design targets. To bridge this gap, initiatives such as the Open Graph Benchmark (OGB)} \cite{hu2020open} or GNNmark \cite{baruah2021gnnmark} \hl{aim to provide a representative set of graphs and GNNs to use as benchmarks. In hardware accelerators, comparing multiple recent architectures is difficult and some works have compared their fundamental dataflows instead} \cite{li2021gcnax}. In this direction, Garg \emph{et al.} \hl{perfomed a dataflow classification that includes multiple operation orders, and whose analysis which may guide further developments in the field} \cite{garg2021taxonomy}.

\hl{A second reflection is that the desirable} \emph{one approach fits all} \hl{does not apply to GNNs, and distinct design approaches will probably be required for different applications. For example, the extreme scale and high throughput demands of recommendation systems is well in line with the targets of software frameworks: programmability and scalability. In contrast, for applications that need to focus on real-time operation and energy efficiency, custom hardware acceleration solutions may be the only way to go. Moreover, the wide variety of problems with their different graph and feature vector sizes renders the acceleration problem more difficult to tackle with a single approach} \cite{EnGN,Yan2020a,Zhang2020b}.

\hl{Finally, we identify a few outstanding challenges for acceleration. Support for dynamic graphs is a pending issue only evaluated in AliGraph} \cite{Zhu2018ali}\hl{. Learning over dynamic graphs implies not only processing the GNN in each time step, but also updating the weight matrices as the graph evolves, factors that might be amenable to software or hardware optimization. At the frontier of software and hardware, another challenge resides in how to approach the GNN acceleration problem with a co-design strategy, i.e. which tasks can be offloaded to software and which ones should stay in hardware, taking into consideration the related overheads. On the hardware side, how to best accelerate training remains as an important open question as all proposals except GraphACT} \cite{Zeng2020} \hl{have targeted inference. Beyond that, another challenge in hardware accelerators is finding the right balance between performance and generalization in light of the multitude of graph types and GNN variants, including techniques such as pooling, sampling, or skip connections.}

\section{GNN Acceleration: The Vision} \label{sec5}
Previous sections have discussed how \acrshort{gnnpl} can be understood as a set of classical NNs working symbiotically over graph-structured data. We have seen that, to extract specific knowledge from the graphs, different NN layers may be employed leading to a wide variety of GNN flavours. This, plus the fundamental dependence of GNNs on the input graph (which may be extremely large) complicate the task of streamlining their execution. As a result, works on GNN acceleration have implicitly made a choice upon either providing an extremely efficient acceleration scheme for a specific GNN variant, or being general or flexible enough to serve multiple types of GNNs less efficiently.

\begin{figure*}
    \centering
    \includegraphics[width=0.9\textwidth]{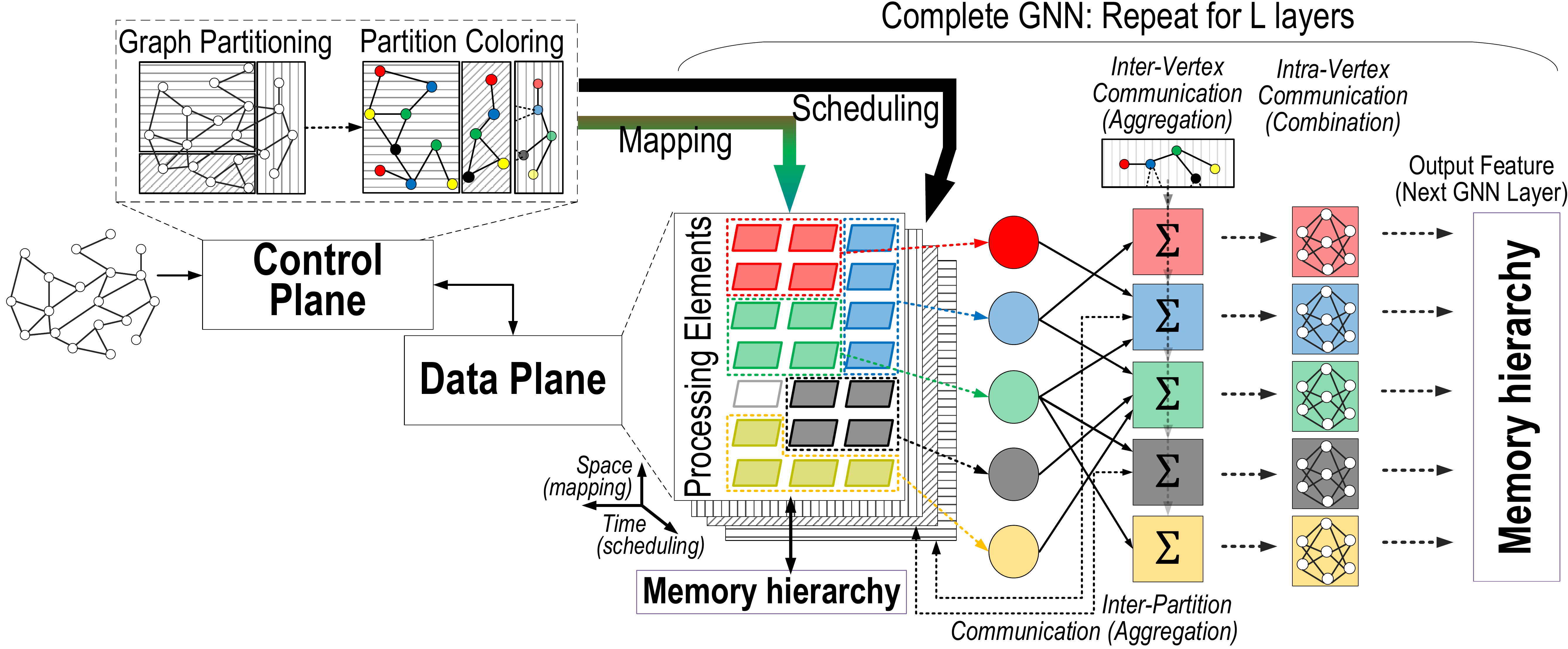}
    \vspace{-0.3cm}
    \caption{Architectural vision for GNN accelerators with hardware-software co-design (i.e. control and data planes), graph awareness (i.e. guided mapping and scheduling), and communication-centric design (i.e. reconfigurable interconnect).}
    \label{Fig4} \vspace{-0.4cm}
\end{figure*}

The key challenge in GNN acceleration is thus to provide a framework that is able to both maximize performance and efficiency while maintaining a degree of flexibility that caters to the different graph sizes, characteristics, and GNN algorithms. Albeit a daunting task, in this section we aim to leverage the analysis of existing acceleration works to hypothesize which would be the main characteristics that future GNN accelerators should feature. In particular, our envisaged architectural approach shall be driven by (i) software-hardware co-design, (ii) graph awareness, and (iii) an much-needed emphasis on communications. We next discuss these aspects qualitatively, using Figure \ref{Fig4} as reference.

\subsection{Software-Hardware Co-Design}
The analysis of prior work has shown that both software and hardware approaches can provide significant speedups. In some occasions, one might argue that both strategies attack the problem similarly, e.g. node reordering in software \cite{Tian2020} and workload balancing in hardware \cite{AWB}. However, a few works have also started to realize that both approaches are not mutually exclusive and that their benefits can add up, or one can simplify the other. For instance, \hl{Rubik improves performance by reordering the graph in software} \cite{chen2020rubik}. Also, the design from Zhang \emph{et al.} \cite{Zhang2020acc} eliminates redundant operations via software pre-processing and then optimizes execution with specialized aggregation and combination modules. The software side allows to avoid having specialized hardware structures to eliminate redundant operations. 

Building upon this observation, our first proposed pillar is \emph{software-hardware co-design} as a strategy for handling different GNNs and graphs efficiently while retaining some hardware simplicity. We advocate for a control-data plane model where, in general, the control plane will be implemented entirely in software providing the flexibility and the data plane will be implemented in custom hardware providing the efficiency. While conceptually separated (see Fig. \ref{Fig4}), the operation of both planes will be tightly coupled.

On the one hand, the \textbf{control plane} manages the actions of the accelerator by having a global view of the complete GNN structure and input graph. The control plane is responsible for dictating the dataflow running in the data plane, by (i) partitioning the \acrshort{gnn} computation into manageable computational segments, (ii) mapping the different vertices and edges to the hardware resources of the data plane, and (iii) scheduling the different executions towards balancing the workload, maximize the benefits of pipelining, and so on. Finally, we also consider part of the control plane to (iv) drive pre-processing (and possibly offline) steps such as the removal of redundant operations \cite{Jia2020} or the detection of certain graph aspects such as cliques \cite{gnnadvisor}. By being implemented in software, all these functions can deliver the required flexibility to accelerate any GNN workload. \hl{However, given that certain pre-processing steps may take minutes or hours in very large graphs} \cite{chiang2019cluster}\hl{, care must be taken in not turning the software side into the bottleneck of the system. To this end, one may resort to lightweight heuristics or limit software techniques to specific cases such as deep GNNs or training, where the result of pre-processing may be reused multiple times.}

On the other hand, the \textbf{data plane} consists of the processing and memory elements that work as per the control plane instructions to execute a GNN. As we have seen in Section \ref{sec:HW}, we could adopt many strategies for architecting the data plane, e.g., unified, phased, modular, homogenenous, heterogeneous, to name a few. However, we find particularly interesting the use of architectures similar to that of MAERI \cite{maeri}, where an homogeneous array of PEs and a specialized memory hierarchy are put together via a lightweight reconfigurable interconnect fabric. This architecture could adapt the dataflow according to the control plane commands, thus allowing to give service to the multiple execution stages of an algorithm or different algorithms.


\subsection{Graph Awareness}
Most accelerators have attempted to provide methods that adapt to runtime conditions while being largely unaware of the input graph characteristics \cite{Jia2019, EnGN}. However, it has been also realized that aspects such as the size of the graph, the relative size of the feature vectors, the clustering factor of the graph, or the average degree of the same can be extremely relevant in accelerating the GNN \cite{Yan2020, Tian2020, fey2019fast}. In fact, GNNAdvisor \cite{gnnadvisor} seeks to exploit this information explicitly to improve the performance in GPUs, while others have based the order of operations or the mapping of PEs on characterstics of the graph \cite{Yan2020, AWB, EnGN}. \hl{Other characterization works have shown that the impact of loop ordering or dataflow design decisions on performance certainly depends on the input graph} \cite{li2021gcnax, baruah2021gnnmark, garg2021taxonomy}.

This leads to the second pillar of our envisaged architecture: \emph{graph awareness}. If the GNN depends on the input graph, then maximizing performance needs to be aware of the main features of that graph. Offline or online methods shall be used to extract useful information from the graph that, in our case, will be leveraged by the control plane. This will thus affect aspects such as the graph partitioning \cite{Tian2020}, which may be more or less aggressive depending on the degree distribution; the ordering and pipelining of the different aggregate--combine phases, which may vary across layers and across graphs; or the scheduling process to minimize inter-partition communication. A good example of this approach is community detection, whose efficient implementation \cite{FORTUNATO201075,MALLIAROS201395} or prediction \cite{Chen2019} may allow for the partition of the graph in densely connected graphlets at runtime. This is relevant to efficient pooling \cite{ying2018hierarchical}, redundancy elimination \cite{Jia2020}, and optimal scheduling \cite{gnnadvisor}. \hl{Again, it is critical to minimize the overhead of techniques providing graph awareness, either via heuristics, reuse of prior analyses, or its use only in certain occasions where the pre-processing can be done in advance or its benefit maximized, i.e. training.}




\subsection{Communication-Centric Design}
Data movement is the enemy of efficient architectures. Hardware accelerators aim to minimize it by adapting its resources to the execution dataflow but, surprisingly, traditional DNN accelerators \cite{shi,Sze2017} have generally given a relatively low importance to the sub-system handling data movement: the interconnect fabric. This is also true for GNN accelerators, which are generally computing-centric with few exceptions \cite{Tripathy2020, Zeng2020}. However, GNNs pose the additional challenge of not having a single optimal dataflow given the input graph dependence and the many algorithm variants. Thus, data movement continues to be a crucial aspect \cite{guirado2021characterizing}. 



For this reason, the third pillar of our envisaged architecture is taking a \emph{communication-centric design} approach. This is a philosophy that has been applied to endow DNN accelerators with certain flexibility \cite{maeri,Kwon2018c, Kwon2017} or to optimize distributed learning \cite{Tang}. In our case, we propose the use of a reconfigurable interconnect fabric among the PEs to adapt the hardware to the underlying graph connectivity or, in other words, to the optimal dataflow that may vary across layers, partitions, or graphs. In an extreme case, one could adopt the approach of recent DNN accelerators that orchestrate all data movement at compilation time \cite{james2020ispd, abts2020think}. GNNs and their extreme size might discourage the use of this strategy and, instead, advocate for a compilation that provides hints for the interconnect to adapt to the varying needs of the graph and its most optimal dataflow.   
The compilation and reconfiguration could be complemented by the analysis of the input graph. \hl{Assuming it can be done in advance or with little overhead,} graph profiling may allow us to predict the prevalent communication patterns and, thus, the most appropriate interconnect topology.

\section{Conclusion} \label{conclusion}
The recent interest in geometric deep learning, or methods able to model and predict graph-structured data, have led to an explosion of research around GNNs. As we have seen in our analysis of the current state of the art, most of the works focus on the algorithms and their applications, rendering the topic of GNN computing a less beaten path. However, we anticipate that the area of software and hardware support for GNNs will grow at a fast pace, continuing an upwards trend that we observed from 2018 to today. 

The reasons for the probable increase in research delving into more efficient computing means for GNNs are several. First, the field is maturing and the more theoretical algorithm-driven research gives way to the most application-oriented development. A clear example of this trend is the advent of efforts to unify aspects such as benchmarking \cite{hu2020open}. Second, GNNs are the key to many disruptive applications in multiple fields, thus creating a clear application pull driving the need for better processing. Third, GNNs present multiple unique challenges such as the wide variety of algorithm variants, their dependence on the graph characteristics, or their massive scale in some applications. This makes the field of GNN processing unlikely to saturate in the foreseeable future and calls for an in-depth discussion of not only the challenges associated to GNN processing, but also of possible ways to tackle them.


Finally, we highlight the rising popularity of software frameworks and the recent appearance of hardware accelerators for GNNs. On the software side, libraries such as DGL or NeuGraph aim to speed up and add features to widespread frameworks such as TF or PyTorch. Interesting contributions are acceleration of GNNs via graph analysis or pre-coding, as well as the distribution of computation in large-scale systems, much needed for huge recommendation systems. On the hardware side, we did not observe a clear architectural trend and existing proposals are debating between being specific or applicable to multiple GNN variants, and between unified architectures or more hierarchical, tiled organizations. Building on this observation, we envision that future accelerators shall adopt a hardware-software co-design approach to maximize performance, keep graph awareness as a profitable optimization opportunity, and tackle workload variability via a reconfigurable interconnect.



\section*{Acknowledgments}
The authors would like to thank the anonymous reviewers and the editorial team for their constructive criticism, which has helped improve the quality of the paper. We are also grateful to Albert Cabellos-Aparicio, Tushar Krishna, Jos\'{e} L. Abell\'{a}n, and Manuel E. Acacio for the countless discussions on the topic. This work is possible thanks to funding from the European Union's Horizon 2020 research and innovation programme under grant agreement No. 863337 (WiPLASH project) and the Spanish Ministry of Economy and Competitiveness under contract TEC2017-90034-C2-1-R (ALLIANCE project) that receives funding from FEDER.





\end{document}